\documentclass[10pt,journal,compsoc]{IEEEtran}
\usepackage{amsmath,amsfonts}
\usepackage{algorithmic}
\usepackage{algorithm}
\usepackage{array}
\usepackage[caption=false,font=normalsize,labelfont=sf,textfont=sf]{subfig}
\usepackage{textcomp}
\usepackage{stfloats}
\usepackage{url}
\usepackage{verbatim}
\usepackage{graphicx}

\def \ie {\emph{i.e.}~}
\def \eg {\emph{e.g.}~}

\def \etal {\emph{et al.}~}
\def \vs {\emph{v.s.}~}

\newcommand{\para}[1]{\noindent \textbf{#1}}

\usepackage{color}
\usepackage{booktabs}
\usepackage{xcolor}
\usepackage{soul}
\soulregister{\cite}7
\soulregister{\ref}7
\soulregister{\para}7
\usepackage{diagbox}
\usepackage{multirow}
\ifCLASSOPTIONcompsoc
  \usepackage[nocompress]{cite}
\else
  \usepackage{cite}
\fi
\begin{document}

\title{Facial Video-based Remote Physiological Measurement via Self-supervised Learning}

% \author{Zijie~Yue\hspace{-1mm}$^{~\orcidlink{0000-0001-1234-1234}}$, 
%         Miaojing Shi\hspace{-1mm}$^{~\orcidlink{0000-0001-1234-1234}}$,~\IEEEmembership{Senior Member,~IEEE,}
%         Shuai~Ding\hspace{-1mm}$^{~\orcidlink{0000-0001-1234-1234}}$,~\IEEEmembership{Member,~IEEE}
\author{Zijie~Yue, 
        Miaojing Shi,~\IEEEmembership{Senior Member,~IEEE,}
        Shuai~Ding,~\IEEEmembership{Member,~IEEE}

% \IEEEcompsocitemizethanks{\IEEEcompsocthanksitem Z.Yue is with the College of Electronic and Information Engineering, Tongji University, China, and also with the School of Management, Hefei University of Technology, China. (e-mail: zijie.yue@kcl.ac.uk).
\IEEEcompsocitemizethanks{\IEEEcompsocthanksitem Z.Yue and M.Shi are with the College of Electronic and Information Engineering, Tongji University, China. (e-mail: zijie.yue.94@gmail.com, mshi@tongji.edu.cn).
\IEEEcompsocthanksitem S.Ding is with the School of Management, Hefei University of Technology, China. (e-mail: dingshuai@hfut.edu.cn).

(Corresponding author: Miaojing Shi.)}}

% The paper headers
\markboth{IEEE Transactions on Pattern Analysis and Machine Intelligence}%
{Shell \MakeLowercase{\textit{et al.}}: A Sample Article Using IEEEtran.cls for IEEE Journals}

\IEEEtitleabstractindextext{
\begin{abstract}
Facial video-based remote physiological measurement aims to estimate remote photoplethysmography (rPPG) signals from human facial videos and then measure multiple vital signs (\eg heart rate, respiration frequency) from rPPG signals.
%\st{and predict multiple vital signs (\eg heart rate, respiration frequency) from human face videos}. 
Recent approaches achieve it by training deep neural networks, which normally require abundant facial videos and synchronously recorded photoplethysmography (PPG) signals for supervision. However, the collection of these annotated corpora is not easy in practice.
% and small-scale datasets limits the robustness and generalization ability of existing data-hungry models. 
In this paper, we introduce a novel frequency-inspired self-supervised framework that learns to estimate rPPG signals from facial videos without the need of ground truth PPG signals.  
%follow state of the art approaches to exploit pulsation clues from face videos for rPPG \miaojing{rPPG not defined} estimation, but we propose a \ZJ{frequency-inspired self-supervised} learning framework for training the estimator without annotations. 
Given a video sample, we first augment it into multiple positive/negative samples which contain similar/dissimilar signal frequencies to the original one. Specifically, positive samples are generated using spatial augmentation; negative samples are generated via a learnable frequency augmentation module, which performs non-linear signal frequency transformation on the input 
%\st{while retains its visual appearance} 
without excessively changing its visual appearance.  
Next, we introduce a local rPPG expert aggregation module to estimate rPPG signals from augmented samples. It encodes complementary pulsation information from different face regions and aggregates them into one rPPG prediction. Finally, we propose a series of frequency-inspired losses, \ie  frequency contrastive loss, frequency ratio consistency loss, and cross-video frequency agreement loss, for the optimization of estimated rPPG signals from multiple augmented video samples. 
%These losses can also guide the model to learn to detect the signal frequency differences between videos thus train it to capture the most important frequency information for rPPG regression.} 
We conduct rPPG-based heart rate, heart rate variability, and respiration frequency estimation on five standard benchmarks. The experimental results demonstrate that our method improves the state of the art by a large margin. Our codes will be available at \emph{https://github.com/yuezijie/Video-based-Remote-Physiological-Measurement-via-Self-supervised-Learning}.
\end{abstract}

}
\maketitle

\IEEEraisesectionheading{\section{Introduction}}
\IEEEPARstart{P}{hysiological} signals, such as heart rate (HR), heart rate variability (HRV) and respiration frequency (RF), are important vital signs to reflect human health status. Medical equipment like electrocardiography (ECG) and photoplethysmography (PPG) recording devices can measure these signals in a skin-contact way. However, skin-contact electrodes and wires often cause inconvenience and discomfort to users, sometimes lead to allergic reactions \cite{niu_rhythmnet_2020,yue_deep_2021}. Recently, there is a growing interest in measuring physiological signals from human facial videos captured by RGB cameras \cite{lee_meta-rppg_2020,lu_dual-gan_2021,chen_video-based_2019}. This remote physiological measurement has been used in a number of applications, such as atrial fibrillation screening \cite{yan_high-throughput_2020,liu_vidaf_2022} and driver status assessment \cite{nowara_sparseppg_2018}.

\begin{figure*}[!t]
  \centerline{\includegraphics[width=6in]{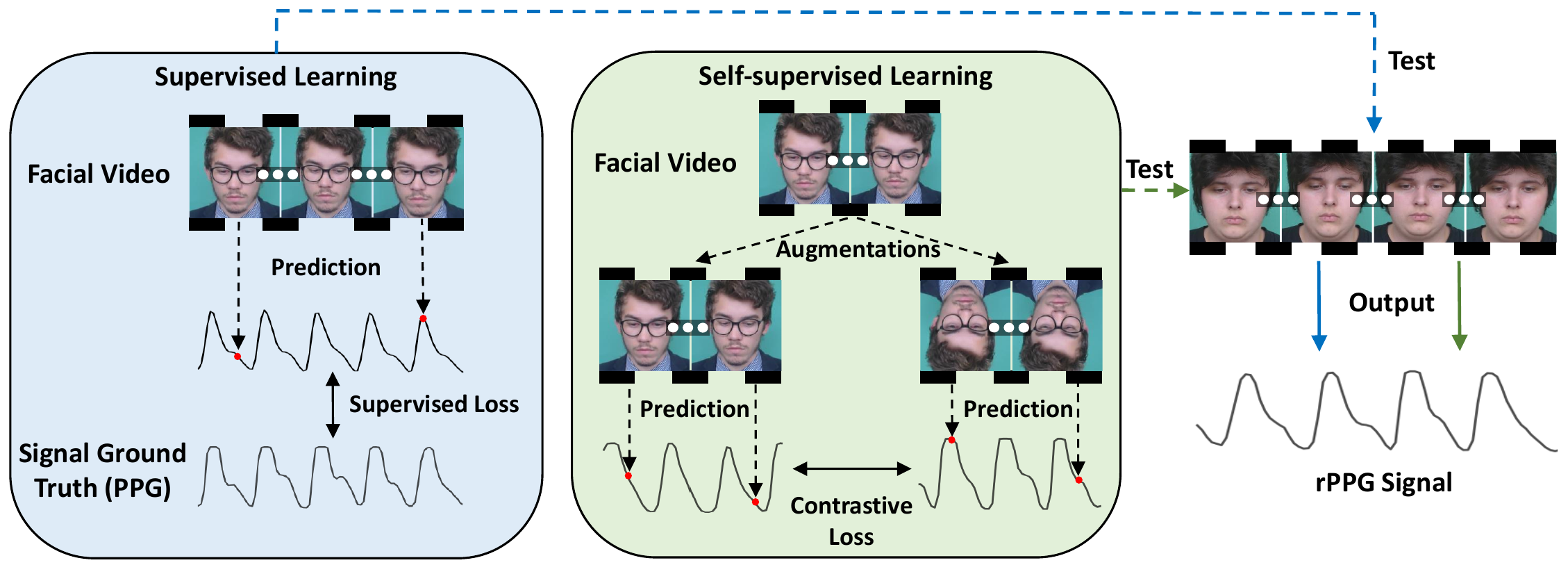}}
  %\vspace{-0.1in}
  \caption{Modern remote physiological measurement approaches train DNNs to estimate rPPG signals from facial videos. Supervised methods require signal ground truth recorded synchronously with facial videos for model training. Based on the proper sample augmentation and contrastive learning, our self-supervised method trains the model from unlabelled facial videos without pre-recorded PPG signals.
  %   Overview of our frequency-inspired self-supervised framework for video-based remote physiological measurement. Two key modules, \ie learnable frequency augmentation and local rPPG expert aggregation, are offered for data augmentation and rPPG estimation, respectively. In the end, a series of frequency-inspired losses are devised among augmented videos with different rPPG signal frequencies and across temporally adjacent videos  with similar rPPG signal frequencies. 
  %The  former  augments the input into multiple negatives with different underlying signal frequencies. The latter estimates rPPG signals from  local face regions and aggregate them into one final prediction. 
  %We apply spatial/frequency augmentations on given video to generate multiple positive/negative samples which contain signals at similar/dissimilar frequencies. Then the frequency differences between them inspire the within-video self-supervised learning. Moreover, based on the observation that the frequency of a rPPG signal does not change rapidly over a short time interval, we also optimize the rPPG signals estimated from positive samples with their temporal neighbors by cross-video self-supervised learning.}
  }
  \label{fig1}
  \vspace{-0.1in}
  \end{figure*}
% \textcolor{red}{}
Most remote physiological measurement approaches are based on the remote photoplethysmography (rPPG) principle \cite{wang_algorithmic_2017}, \ie the optical absorption of skins changes periodically along with the periodic change of blood volume. Ideally, the change of skin color over time reflects the periodic rPPG signal and can be used to measure physiological signs including HR, HRV and RF. Nevertheless, periodic rPPG signals are easily affected by non-periodic noises, which can be caused by illumination changes, facial expressions and head movements~\cite{lu_dual-gan_2021,yue_multimodal_2022}. Traditional approaches propose blind signal separation techniques \cite{mcduff_camera_2023,poh_advancements_2011,poh_non-contact_2010} and skin reflection models \cite{wang_algorithmic_2017,de_haan_robust_2013} to separate rPPG signals from noises. These approaches rely on prior assumptions, \eg the normalized skin tone is the same for different people under white light \cite{de_haan_robust_2013}, which can not always be satisfied, hence leading to performance degradation~\cite{song_pulsegan_2021,niu_video-based_2020}. 
%However, the noises are more complicated, diverse and unpredictable in the real scenarios, \ZJ{\eg the illumination condition varies greatly from indoors to outdoors, and people usually do kinds of unpredictable facial expressions besides nose wrinkler, lip stretcher and brow lowerer.} Many prior assumptions which the traditional approaches rely on do not meet at all, leading to their severe performance decrease .

With the advent of deep learning \cite{he_deep_2016}, modern remote physiological measurement approaches train deep neural networks (DNNs) to estimate rPPG signals from facial videos \cite{niu_rhythmnet_2020,lee_meta-rppg_2020,yue_deep_2021,song_pulsegan_2021,niu_video-based_2020,lu_dual-gan_2021,yu_remote_2019,yu_physformer_2022}. 
%They normally consists of a video preprocessing phase and a feature extraction phase. 
%In the first phase, they locate facial regions of interest (ROI) from videos, then construct various hand-crafted feature representations (\eg spatial-temporal maps, time-frequency maps) of ROI sequences . \ZJ{In the second phase, they} map these representations to target rPPG signals via deep neural networks. 
These methods are trained in a supervised manner, where PPG signals are recorded synchronously with facial videos for supervision (see Fig.~\ref{fig1}). The collection of these annotated corpora is however not easy: subjects must wear skin-contact devices and keep body still to record both signals and videos. To bypass this process, 
%Thus, it is very difficult to collect large-scale datasets for training existing data-hungry models \cite{tsou_multi-task_2021}. 
Gideon~\etal~\cite{gideon_way_2021} have tried to train an rPPG estimator based on unlabeled data using contrastive learning. Contrastive learning normally treats multi-views of a given sample as positive samples, and other samples in the dataset as negative samples. Sample representations are optimized by pulling positive samples close and negative samples away in the feature embedding space. In this context,
% They define each face video and its estimated rPPG signal as the anchor sample/signal. Then they resample the anchor sample to a negative sample and measure the corresponding negative signal. Finally they transform the negative signal to the positive signal according to a frequency ratio, and leverage the contrastive loss for model training.
given an input video, Gideon~\etal~re-sample it with a pre-defined ratio to its negative counterpart, which has a different rPPG signal frequency. % yet retains mostly the spatial information. %which \st{has similar spatial information as the anchor sample, but} contains an underlying rPPG signal with another frequency. 
%The input video is taken as an anchor sample, iteration, they \st{define a face video as an anchor sample. Then, based on a resampling ratio, they} \ZJ{first augment a face video (anchor sample)} \st{the anchor sample} to a negative sample, 
Positive and negative rPPG signals are respectively estimated from both the input and resampled videos. 
%The latter one is also reversely sampled to create the positive sample. 
The contrastive loss is applied among these signals for model optimization. \cite{gideon_way_2021} is the first DNN-based self-supervised work for rPPG estimation. 
But it generates only one negative sample with a higher rPPG signal frequency to that of the input, which limits the sample diversity and can lead to inferior model generalizability. 

In this paper, we propose a novel frequency-inspired self-supervised framework for facial video-based remote physiological measurement, which learns to optimize rPPG estimation from multiple augmented videos of different signal frequencies and across temporally neighboring videos of similar signal frequencies.
% (see Fig.~\ref{fig1}) 
%Three key components are offered to enable this framework (see Fig.~\ref{fig1}b): the learnable frequency augmentation module that learns to augment input video into multiple videos with different  which have different frequencies of rPPG signals, region-level experts aggregation module (REA) and multiple new loss functions defined in frequency domain.}
It has three main stages: data augmentation, signal extraction and network optimization. Given an input video, \emph{in the first stage}, we apply spatial augmentation (\eg image rotation and flip) to obtain multiple positive samples, which contain rPPG signals with the same frequency to that of the input. 
%Then considering that the most important goal of the rPPG estimator is to measure rPPG signals at various frequencies, 
Meanwhile, we design a learnable frequency augmentation (LFA) module to generate multiple negative samples which contain rPPG signals with different frequencies
to that of the input. Unlike~\cite{gideon_way_2021}, our LFA module is integrated into the DNN for end-to-end training and produces multiple samples with controllable rPPG signal frequencies, which can be higher or lower than that of the input. \emph{In the second stage}, because face is not an ideal Lambertian object, the distribution of blood vessels or noises varies over different regions in a face. Instead of treating face regions indiscriminately \cite{yue_deep_2021,gideon_way_2021}, we design a local rPPG expert aggregation (REA) module to estimate rPPG signals from different face regions and aggregate them via
%We associate each local rPPG expert is associated with an attention block to focus the estimation on its pulsation-sensitive area.
a spatio-temporal gating net. The REA module is respectively devised for each augmented video sample (positive or negative) to obtain its rPPG signal. 
%\ZJ{thus these regions have mutual and their exclusive rPPG clues} \cite{lu_dual-gan_2021,favilla_heart_2019},
\emph{In the third stage}, since we do not have ground truth PPG signals, we first introduce a frequency contrastive loss and a frequency ratio consistency loss to optimize our estimated rPPG signals. Furthermore, since the frequency of rPPG signal does not change rapidly over a short term, we additionally extract rPPG signals from multiple temporally neighboring videos to the input and introduce a cross-video frequency agreement loss for rPPG estimation across videos. 
%\st{This loss improves the periodicity of the estimated rPPG signal} 

%\ZJ{This loss constrains the frequency conformability of the estimated rPPG signal with other adjacent signals to reduce frequency estimation errors and improve signal periodicity.}

%while pushing them away from the negative signals using a frequency contrastive loss. 
%Further, we introduce two new losses We also introduce a frequency ratio consistency loss to enhance the frequency augmentation ability of LFAM. 

%. \ZJ{We carefully ensure the frequency similarities between positive samples and LFAM learns to expand the frequency dissimilarities between positive samples with negative samples for more effective self-supervised learning.}

% Its frequency augmentation operations are performed on the multi-scale features through a Pyramid, Cascading and Modulation structure. 

Overall, our new frequency-inspired self-supervised framework for remote physiological measurement has three key contributions:

\begin{enumerate}
   \item We design a learnable frequency augmentation module to generate sufficient and diverse negative samples for self-supervised learning. 
   %It performs frequency augmentation on the multi-scale features of anchor sample through its Pyramid, Modulation and Aggregation structure.
   \item We introduce a local rPPG expert aggregation module to learn complementary information from different face regions for rPPG estimation. %We design RA-Experts to automatically assign higher attentions to more informative skin areas for better estimation performance.
   \item We propose a series of frequency-inspired losses (\ie frequency contrastive loss, frequency ratio consistency loss, cross-video frequency agreement loss) for the optimization of rPPG signals extracted from the input's augmentations and neighbors. 
   %We also design a frequency ratio consistency loss and a cross-video frequency agreement loss to improve the frequency augmentation effects of LFAM and enhance the periodicity of rPPG signals.
\end{enumerate}

We conduct extensive experiments on five standard benchmarks, \ie UBFC-rPPG \cite{bobbia_unsupervised_2019}, PURE \cite{stricker_non-contact_2014}, DEAP \cite{koelstra_deap_2012}, MMVS \cite{yue_deep_2021} and BP4D+ \cite{zhang_multimodal_2016}. Results show our approach significantly outperforms the state of the art self-supervised method and performs on par with state of the art supervised methods.

\section{Related work}

\subsection{Remote physiological measurement}
Facial video-based remote physiological measurement 
%aims to measure heart rate, heart rate variability and respiration frequency via rPPG signals estimated from face videos \cite{chen_video-based_2019,sun_photoplethysmography_2016}. It 
has been mainly implemented using blind signal separation \cite{poh_advancements_2011,macwan_heart_2019,mcduff_fusing_2018}, skin reflection models \cite{wang_algorithmic_2017,de_haan_robust_2013}, and deep neural networks~\cite{niu_rhythmnet_2020,lee_meta-rppg_2020,song_pulsegan_2021}. First, blind signal separation-based approaches assume that the {skin color change} is a linear combination of the target rPPG signal and other noises. Signal decomposition methods, such as independent component analysis (ICA) or principal components analysis (PCA), are used to separate rPPG signals from noises. For example, Macwan \etal \cite{macwan_heart_2019} used auto-correlation as a measurement for signal periodicity, so as to guide ICA for rPPG signal separation. 
McDuff \emph{et al.} \cite{mcduff_fusing_2018} leveraged ICA to estimate rPPG signals from multiple facial videos captured at different angles.
Second, skin reflection model-based approaches explore different ways to project images from the RGB space to other color spaces for better rPPG estimation. For instance, Haan \etal \cite{de_haan_robust_2013} proposed to project images into the chrominance space to eliminate motion noises and help rPPG estimation. These approaches rely on assumptions, for instance, different people's skin tones are identical under white light \cite{de_haan_robust_2013}, which may not always be satisfied in a practical environment \cite{song_pulsegan_2021,niu_video-based_2020}. We refer readers to \cite{mcduff_camera_2023} for a comprehensive and systematic survey on these traditional approaches.

Recently, DNN-based approaches have shown superior performance on remote physiological measurement \cite{niu_rhythmnet_2020,lee_meta-rppg_2020,yue_deep_2021,song_pulsegan_2021,niu_video-based_2020,lu_dual-gan_2021,yu_remote_2019,yu_physformer_2022}.
%Niu \etal \cite{niu_rhythmnet_2020} constructed several hand-crafted spatiotemporal maps to represent face videos and fed them to {several convolution layers and one gated recurrent unit} for heart rate estimation. 
Hu \etal \cite{hu_eta-rppgnet_2021} designed a time-domain attention network to extract and aggregate temporal information from multiple video segments for rPPG estimation. Song \etal \cite{song_pulsegan_2021} transformed the chrominance signals in~\cite{de_haan_robust_2013} into accurate rPPG waveforms using a conditional generative adversarial network. Yu \etal \cite{yu_remote_2019} and Yue \etal \cite{yue_deep_2021} respectively addressed rPPG estimation from highly compressed and low-resolution facial videos by adding video enhancement networks. 
%Niu \etal \cite{niu_video-based_2020} proposed a multi-task network to estimate rPPG and heart rate simultaneously. 
%and introduced a cross-verified feature disentangling strategy to separate target physiological information and other noises. 
Yu \etal \cite{yu_physformer_2022} used transformer blocks to model the relationship among video frame features. 
Despite {their} remarkable achievements, these approaches need plenty of facial videos with synchronously recorded PPG signals for model training. The collection of these annotated corpora is however not easy. Gideon \etal \cite{gideon_way_2021} tackled this by training the DNN in a self-supervised manner using unlabelled data: they resampled the input video to create its negative counterpart and applied a contrastive loss among rPPG signals extracted from the input and resampled videos.
% Given anchor videos, they manually resample them to the negative videos which contain underlying rPPG signals at different frequencies, and then force their model to learn to determine the frequency differences for estimating the signal of interest. 
% Our work is similar to them that we both use contrastive learning architecture to train the rPPG estimator.

We propose a new frequency-inspired self-supervised framework for facial video-based remote physiological measurement. It differs from \cite{gideon_way_2021} significantly:  
1) for data augmentation, \cite{gideon_way_2021} uses pre-processed video re-sampling to generate one negative sample, whose rPPG signal frequency can only be higher than that of the original sample. Our LEA module is instead learned to augment samples with arbitrary signal frequencies; 
%Moreover, the positive rPPG signal in \cite{gideon_way_2021} is obtained by inverse transformation from the negative one while we generate them separately. 
%Our strategy significantly improves the frequency diversity of augmented samples, thus benefits to the model generalizability;  
%and the manual resampling operation tend to generate bad quality samples \st{facing} \ZJ{when inputting} low-frame-rate videos. %While we propose learnable LFAM to arbitrarily increase or decrease the frequencies of underlying rPPG signals through its non-linear frequency modulation operations. 
%\st{Moreover, the positive signal in} 
% \cite{gideon_way_2021} 
%\st{is directly transformed from the negative signal rather than generating from positive samples, leading to severe frequency inconsistency between it and the anchor signal. Compulsively minimizing their distance makes the estimator hard to converge. While we instead obtain positive signal from positive samples to avoid this problem.}
2) for signal extraction, \cite{gideon_way_2021} only captures image-level features and treats different face regions indiscriminately. Our REA module instead considers the  {pulsatile feature} variance over face regions and advantageously aggregates their information for rPPG estimation; 
% considering that both the vascular and noise distributions are totally different in these regions \cite{lu_dual-gan_2021,favilla_heart_2019}, 
%we encode complementary information from different regions via RA-experts accordingly, and use a temporal gating network to combine them for information aggregation. 
3) for model optimization, \cite{gideon_way_2021} applies the contrastive loss among anchor, positive and negative rPPG signals. They all come from one video clip.  We design a series of frequency-inspired losses; especially, the cross-video frequency agreement loss is proposed among rPPG signals extracted from multiple video clips.
%\cite{gideon_way_2021} fail to maintain long-range signal frequency consistency. 

% \cite{gideon_way_2021} only generates one negative sample for contrastive learning in each iteration. While we generate multiple negative samples instead, thus improving the model convergence speed and model generalization ability. Moreover, 
% we design a novel frequency contrastive loss, a frequency ratio loss and a projection loss to improve the model performance.

It is worth noting that the three key contributions of our self-supervised framework are also unique among supervised methods in the remote physiological measurement. 

\subsection{Self-supervised learning}

Self-supervised learning aims to learn an effective data embedding function from unlabelled data during training. Previous works focus on designing different pretext tasks to train the data encoders. For image encoders, solving jigsaw puzzles \cite{leibe_unsupervised_2016}, rotation prediction \cite{gidaris_unsupervised_2018} and counting primitives \cite{noroozi_representation_2017} are widely used tasks. For video encoders, predicting frames \cite{vondrick_anticipating_2016} and tracking patches \cite{wang_unsupervised_2015} can help them learn useful feature representations. 
%For instance, based on the image inpainting task, Pathak \etal \cite{pathak_context_2016} trained the context encoder by predicting the missing image patches using other available regions. 
Recently, contrastive self-supervised learning attracts
much attention and shows promising performance compared to supervised learning in both classification-based \cite{chen_simple_2020,he_momentum_2020,pan_videomoco_2021,chen_exploring_2021,wang_contrastive_2022,qian_spatiotemporal_2021} and regression-based tasks \cite{spurr_self-supervised_2021,wang_contrastive_2022_1,dangovski_equivariant_2022,tukra_stereo_2022}.
%\ZJ{It has made breakthroughs in both classification and regression tasks, such as image classification \cite{chen_simple_2020}, semantic segmentation \cite{he_momentum_2020}, and action recognition \cite{pan_videomoco_2021} for classification tasks, hand pose estimation \cite{spurr_self-supervised_2021}, gaze estimation \cite{wang_contrastive_2022_1}, and density-of-states prediction \cite{dangovski_equivariant_2022} for regression tasks. }
Owing to the versatile data augmentation techniques, contrastive learning approaches augment the input into multiple positive counterparts, 
% \miaojing{there are also approaches of masked images, do they belong to views?}
%\st{different views} 
%\ZJ{data similar to} \st{of} an anchor sample as positives,
while considering other samples in the dataset as negatives to the input. They minimize feature distances between positive sample pairs while maximizing feature distances between negative sample pairs. 
%which describe other objects or events. 
%Plenty of positive and negative samples are leveraged to discover the intra-class similarity and inter-class dissimilarity. 
For instance, SimCLR \cite{chen_simple_2020} leverages image cropping, color distortions and blurring to augment the input and applies contrastive learning. Dangovski \etal \cite{dangovski_equivariant_2022} proposed the equivariant contrastive learning, which utilizes an additional branch to predict the adopted augmentations of positive samples. %They inject the equivariant property into the encoder for better feature representation.}
%then after performing ResNet-50 \cite{he_deep_2016} for image feature extraction, it attracts the positive representations and repulses them with negative ones using the contrastive loss. 
Qian \etal \cite{qian_spatiotemporal_2021} designed a novel temporal augmentation method for video representation learning, which samples non-overlapping positive video clips from the original video using different temporal intervals. 
%were {sampled from a monotonically decreasing distribution. 
Pan \etal \cite{pan_videomoco_2021} introduced another temporal augmentation method which obtains positive videos using a generate adversarial network. 
The network removes several frames from a given video while maintaining its spatio-temporal information.
%The values of intervals are sampled under a monotonically decreasing distribution to improve the augmentation effect. 
We realize the self-supervised video-based remote physiological measurement as a regression problem. Unlike the above approaches that use spatial or temporal augmentation, we propose a learnable frequency augmentation module to augment samples that contain different frequencies of rPPG signals.

\begin{figure*}[!t]
  \centerline{\includegraphics[width=6.5in]{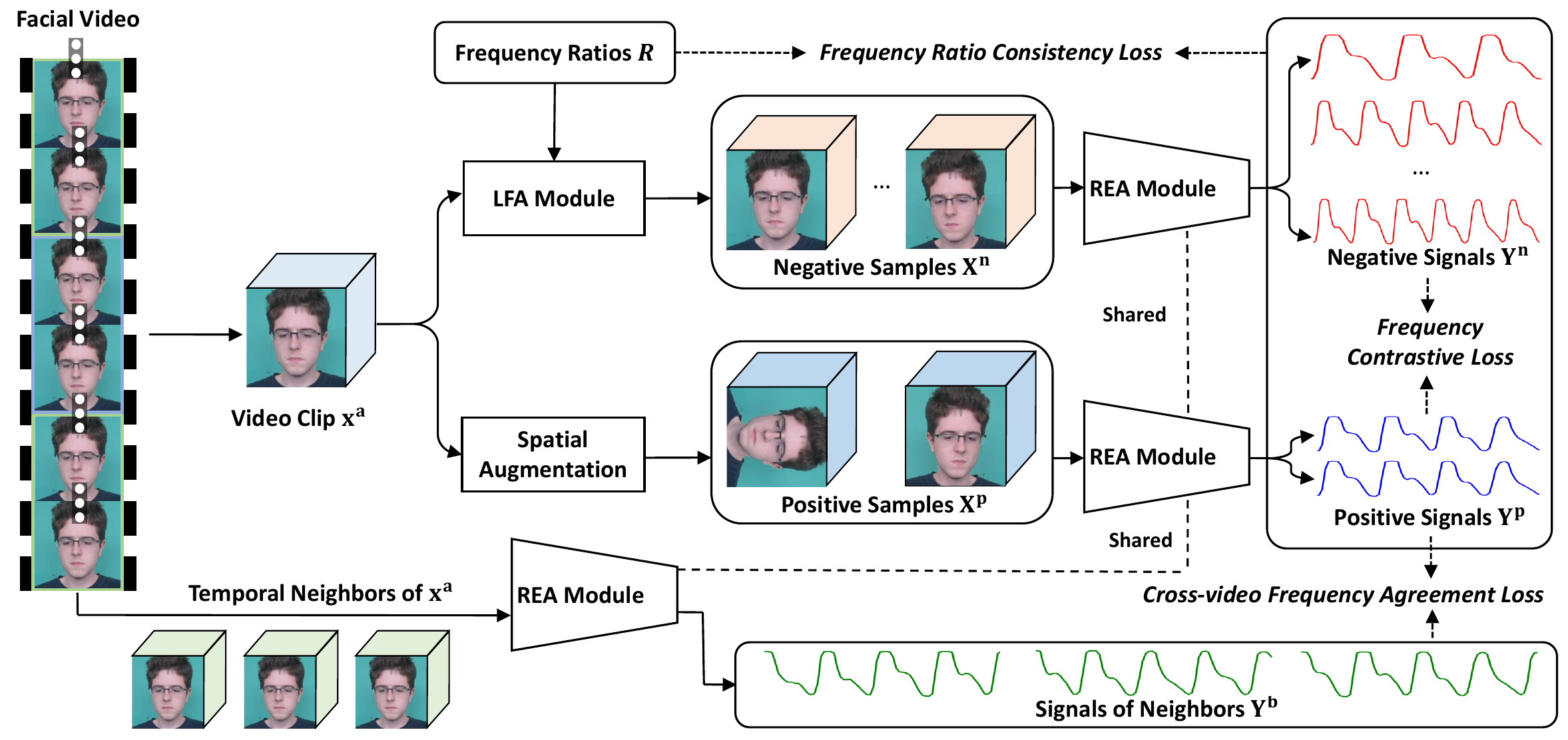}}
  %\vspace{-0.1in}
  \caption{The overall architecture of our framework. It comprises three main stages. 1) Data augmentation: we randomly select a video clip from a short facial video as input while keeping other clips as the input's neighbors for later usage. We respectively apply spatial augmentation and learnable frequency augmentation (LFA)  on the input to generate positive and negative video samples. 2) Signal extraction: we design the local rPPG expert aggregation (REA) module to extract rPPG signals from positive/negative video samples. 3)  Model optimization: we develop a series of frequency-inspired losses, \ie frequency contrastive loss, frequency ratio consistency loss, cross-video frequency agreement loss, for the optimization of rPPG signals generated from the input video's augmentations and temporal neighbors.}
  \label{fig2}
  \vspace{-0.1in}
  \end{figure*}

\subsection{Mixture of experts}

Traditional mixture of experts (MoE) divides the input space into multiple sub-spaces and distributes them amongst multiple experts \cite{jacobs_adaptive_1991,srivastava_data_1999}. A gating net is devised to assign soft or hard weights to experts to control their activations~\cite{jacobs_adaptive_1991}. \emph{Zero coefficient} problem usually occurs in traditional MoE \cite{masoudnia_mixture_2014}: because of the competition among experts, the gating net assigns near-zero weights to some of them due to their unfavourable initial parameters, which may lead to their elimination during training \cite{hansen_combining_1999}. Moreover, the number of samples required for training traditional MoE scales up with the number of experts. Many modern MoE methods tend to extract information from different parts of a sample and combine their predictions via a gating net \cite{yuksel_twenty_2012}. For example, Abdulhamit \emph{et al.}\cite{subasi_eeg_2007} decomposed a given EEG signal into multiple frequency sub-bands using discrete wavelet transform; then predicted the probability of epilepsy from each frequency sub-band using a neural network; the outputs of different neural networks were combined via a gating net for the final diagnosis.

%the space partitioning manner in traditional MoE is based on expert performance on different sub-spaces, which originates from different initial weights. This partitioning manner may leads to complex and nested partitions, thus the gating network cannot control experts well. Moreover, 

% {the results of these experts are softly or hardly weighted by gating networks}

MoE has shown superior performance in many fields, such as image super-resolution \cite{emad_moesr_2022}, multi-modal/task learning \cite{shi2019variational,ma_modeling_2018}, and medical imaging~\cite{sahasrabudhe_deep_2021}. For instance, Ma \etal \cite{ma_modeling_2018} proposed a multi-gate MoE for multi-task learning. Experts capture task-shared representations while gating nets combine them using task-specific weights. 
{Sahasrabudhe~\etal~\cite{sahasrabudhe_deep_2021} extracted information from blood smear images and patient clinical attributes (\eg age and lymphocyte count) using a CNN and MLP model, respectively; they combined the outputs of two models via a gating net for the diagnosis of lymphocytosis.}
% \etal 
%\ZJ{To tackle the problem of domain generalizable (DG) ReID, Dai \etal \cite{dai_generalizable_2021} applied several domain-specific experts on source datasets (domains) to exploit their discriminative and complementary information. Then a gating net calculated the unseen target domain’s relevance w.r.t. all source domains for adaptively integrating source experts’ features into the aggregated features, which would have strong generalizability on the target domain.}
%Considering the varying distributions of {blood vessels} and noise in different face regions \cite{lu_dual-gan_2021}, 
In this paper, we for the first time leverage multiple experts to estimate rPPG signals in different face regions and aggregate their complementary information via a spatio-temporal gating net. %\ZJ{This process is achieved by our novel region-level experts aggregation module (REA).} 

\section{Method}

\subsection{Overview}\label{sec3.1}

Our goal is to learn rPPG signals from unlabelled facial videos. The overview of our framework is illustrated in Fig.~\ref{fig2}. The basic pipeline is to first generate positive and negative samples from the input video, then extract rPPG signals from them, and finally learn contrastive frequency information among these rPPG signals.
%\st{to train its spatial-attention MoE-based rPPG estimator (MORE)} 
%through its self-supervised learning pipeline. 
%It contains three main stages: data augmentation,  
%by a spatial augmentation module and a learnable frequency augmentation module (LFAM), signal extraction, and network optimization. 
Specifically, given a short facial video, we first follow~\cite{yu_physformer_2022,li_learning_2023,yu_physformer_2023} to use the open-source face detector MTCNN \cite{zhang_joint_2016} to detect, align and crop the face area in each frame in advance. The face alignment operation in MTCNN aligns the upside-down faces, rotated faces, and even side faces to canonical frontal faces. Then we cut the aligned facial video into several clips, each with $T$ frames. One clip is randomly selected as the main input $\mathbf {x^a}$ while the rest is taken as $\mathbf {x^a}$'s temporal neighbors for later usage. In the data augmentation stage (Sec.~\ref{sec3.2}), we first apply spatial augmentation to $\mathbf {x^a}$, which does not affect its interior rPPG signals, to generate a number of positive samples $\mathbf {X^p}=\{\mathbf{x^p}\}$. Next, we introduce the learnable frequency augmentation (LFA) module to generate a number of negative samples $\mathbf{X^n}=\{\mathbf{x^n}\}$ which have different rPPG signal frequencies to that of $\mathbf{x^a}$. We feed a set of frequency ratios $R = \{r\}$ into the LFA module to modulate the rPPG signal of $\mathbf{x^a}$.  
%Using $F$ as input together with $x_a$, we generate $k$ negative samples $X_n=\left \{x_n^1,x_n^2,\dots,x_n^k\right \}$ through \ZJ{the learnable frequency augmentation (LFA) module} (Sec. 3.2), these negative samples contain undelying rPPG signals at various frequencies than that contained in $X_p$. 
%LFA module transforms the signal frequency on multi-scale features of given video using a Pyramid, Modulation and Aggregation structure (Fig.~\ref{fig3}). 
Having $\mathbf{X^p}$ and $\mathbf{X^n}$, in the signal extraction stage (Sec.~\ref{sec3.3}), we pass them into local rPPG expert aggregation (REA) modules to estimate corresponding rPPG signals, denoted by $\mathbf{Y^p}=\left \{\mathbf{y^p} \right\}$ and $\mathbf{Y^n}=\left \{\mathbf{y^n} \right \}$ respectively. The REA module encodes complementary pulsation information from different face regions and uses a spatio-temporal gating net to aggregate the information for a final rPPG prediction. Last, in the network optimization stage (Sec.~\ref{sec3.4}), the frequency contrastive loss is applied among signals from $\mathbf{Y^p}$ or between $\mathbf{Y^p}$ and $\mathbf{Y^n}$. The frequency ratio consistency loss is applied between signals from $\mathbf{Y^p}$ and $\mathbf{Y^n}$ to enforce their frequency ratios according to input ratios in $R$.  
%between $Y_p$ and $Y_n$ for self-supervised training. 
We also introduce a cross-video frequency agreement loss by extracting rPPG signals from $\mathbf{x^a}$'s temporal neighbors and enforcing their frequencies to be similar to that of $\mathbf{x^a}$. 
%This can help correct frequency errors thus improving signal periodicity (Sec. 3.4).
%from multiple negative and positive samples via the region-level experts aggregation module (REA), and 
  
%\ZJ{using multiple novel loss functions defined in frequency domain} \st{a reconstruction loss, a frequency contrastive loss, a frequency ratio loss and a projection loss}.

% to generate multiple negative samples containing the underlying rPPG signal at different frequencies from the positive samples using FAG, to respectively estimate rPPG signals from multiple samples via MOR, and to train the end-to-end contrastive model in an unsupervised manner based on four loss functions.

\subsection{Data augmentation: learnable frequency augmentation }\label{sec3.2}
Given the input video $\mathbf{x^a}$, we apply two different data augmentation strategies to generate positive and negative samples, respectively. 
For positive ones, we employ six spatial augmentation operations, \ie image rotations (0°, 90°, 180°, and 270°), horizontal and vertical flips. They do not affect the interior rPPG signal of $\mathbf{x^a}$. 
Every time, we randomly select two operations to apply to $\mathbf{x^a}$ and obtain two positive samples $\mathbf{X^p}=\left \{\mathbf{x}^\mathbf{p}_1,\mathbf{x}^\mathbf{p}_2 \right\}$. They have the same rPPG signal frequency to that (${f^a}$) of $\mathbf{x^a}$. 
%For the data augmentation of positive samples, 
%to generate two positive samples:  to $x_a$. Note that the spatial augmentations only include image rotation, image horizontal flip and image vertical flip operations \cite{chen_simple_2020}.
%\ZJ{we first frame-by-frame apply image spatial augmentations to $x_a$ twice thus transform it into two positive samples: $X_p=\left \{x_p^1,x_p^2 \right\}$. Note that the spatial augmentations only include image rotation, image horizontal flip and image vertical flip operations \cite{chen_simple_2020}. The probabilities of performing these three kinds of augmentations are both 50\%. Before generating each positive sample, we consecutively determine whether these augmentations will be applied, and once the augmentation types are determined, frames of $x_a$ will be preformed with the same augmentations.} %These weak augmentations do not bring in additional variations to the temporal signals contained $X_a$.

\begin{figure}[!t]
  \centerline{\includegraphics[width=3.6in]{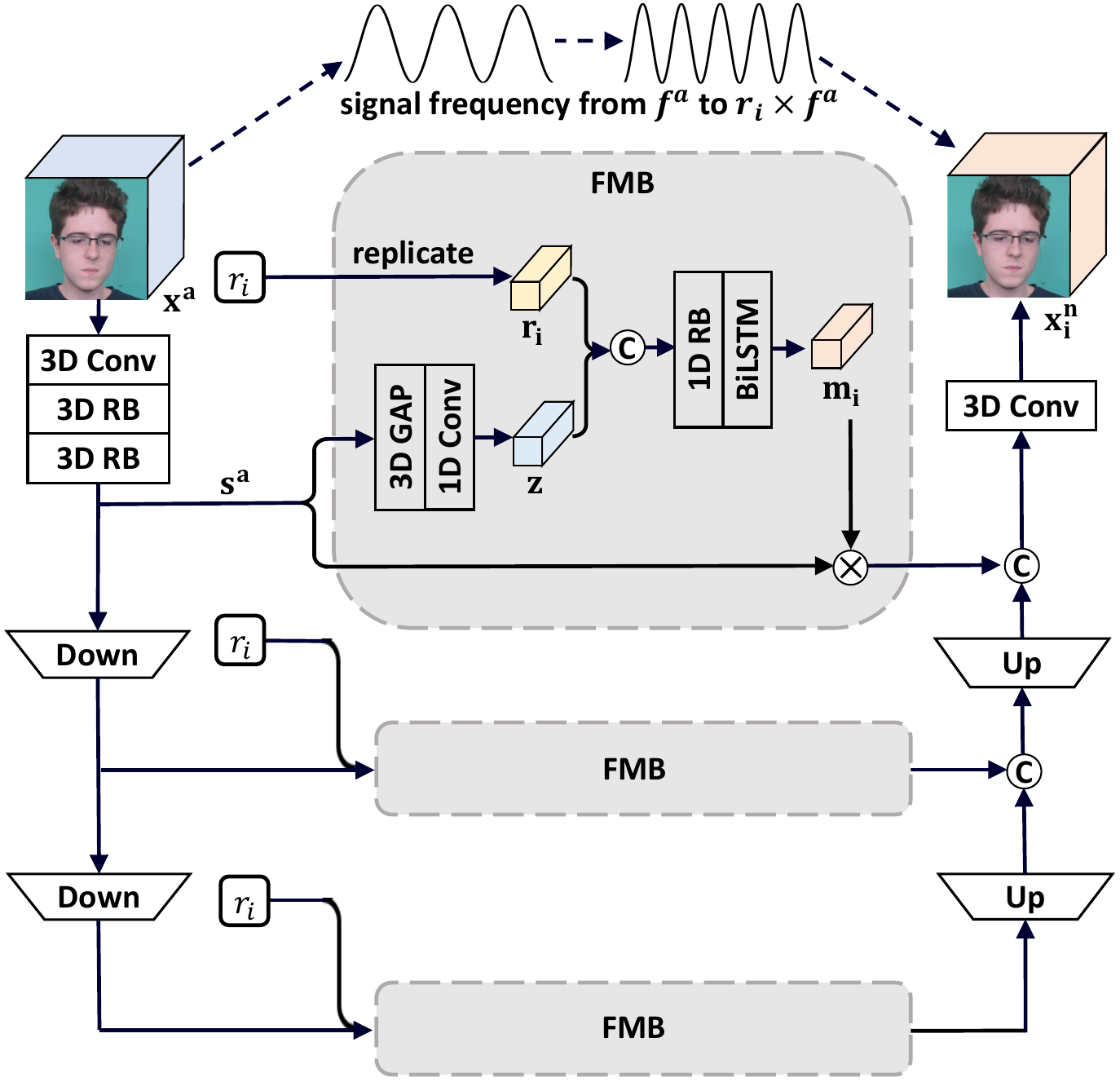}}
  %\vspace{-0.1in}
  \caption{Illustration of the LFA module. It performs frequency modulation on multi-scale features of the input video $\mathbf{x^a}$. The core of LFA module is the frequency modulation block (FMB), which produces the frequency modulation vector given the input scale feature $\mathbf{s^a}$ and the frequency modulation ratio $r_i$. The frequency modulation vector is multiplied back to $\mathbf{s^a}$ to change its rPPG signal frequency. Finally we aggregate multi-scale outputs to obtain the negative video sample $\mathbf{x}^\mathbf{n}_i$.}
  \label{fig3}
  \vspace{-0.1in}
  \end{figure}
  
% As discussed in popular contrastive learning structures}
% \cite{he_momentum_2020,chen_exploring_2021,chen_simple_2020,diba_vi2clr_2021},
% contrastive training needs plenty of negative samples to enforce the encoder to learn to detect whether they have similar or dissimilar representations with positive samples. In our case we need}
Next, we design a learnable frequency augmentation (LFA) module to generate multiple negative samples which contain rPPG signals with different frequencies to that of $\mathbf{x^a}$. Given the input of $\mathbf{x^a}$ and a frequency ratio $r_i$, it has a \emph{pyramid} structure (see Fig.~\ref{fig3}) to transform $\mathbf{x^a}$ to a new sample $\mathbf{x}^\mathbf{n}_i$ with its rPPG signal frequency becoming $r_i \times f^a$. 
$r_i$ is a scalar randomly sampled from a predefined range, which represents the target frequency ratio upon which LFA  modulates the original signal frequency.
%For example, supposedly that the sampled $r_i$ is 0.5, it means the LFA should learn to modulate the signal frequency $f^a$ of $\mathbf{x^a}$ to $0.5\times f^a$ to obtain the negative counterpart $\mathbf{x}^\mathbf{n}_i$.} 
In the \emph{pyramid} structure, we first use one 3D Convolution (Conv) layer followed by two 3D Res-blocks (3D RB) to extract the feature for $\mathbf{x^a}$ and then down-sample it with a factor of 2 and 4 to obtain the multi-scale features.
% ; they are further processed with one more 3D RB, respectively. 
These features reflect different levels of details for $\mathbf{x^a}$. We modulate the frequency of rPPG signal on each scale via the frequency modulation block (FMB, see below Sec.~\ref{sec3.2.1}). The modulated features are upsampled and concatenated. We reconstruct the video clip from the concatenated feature via one 3D Conv ($1\times 1\times 1$). The reconstructed video clip is denoted by $\mathbf x^\mathbf{n}_i$ and is regarded as a negative sample to $\mathbf {x^a}$.

We know the frequency of human rPPG signal varies within a rather small range as a result of the human heart rate limit. On the one hand, we need to reinforce the module to change the rPPG signal frequency of $\mathbf {x^a}$ to the target. On the other hand, we also make sure there is no excessive color difference between $\mathbf {x^a}$ and $\mathbf x^\mathbf{n}_i$,
%{avoid significant frequency chang.} 
because, in reality, the face color changes caused by the blood volume variations are rather subtle.
%\st{retains the general visual appearance of} $\mathbf{x^a}$ \ZJ{because the frequency of human rPPG signal lies in a very small range}. 
We develop the frequency ratio consistency loss (specified later in Sec.~\ref{sec3.4.2}) and video reconstruction loss (see below Sec.~\ref{sec3.2.2}) for the two purposes, respectively. 
Overall, during the training process, we randomly sample a set of frequency ratios $R = \{r_1,..., r_k\}$ within a range of $(0.3,0.8)\cup (1.2,1.7)$ and feed them into the LFA module to augment $k$ negative samples $\mathbf{X^n}=\{\mathbf{x}^\mathbf{n}_1,...,\mathbf{x}^\mathbf{n}_k\}$ to $\mathbf{x^a}$, which should contain rPPG signals with frequencies of $\{r_1\times f^a,...,r_k\times f^a\}$.

\subsubsection{Frequency modulation block}\label{sec3.2.1}
The FMB, as illustrated in Fig.~\ref{fig3}, takes the input of a certain scale feature $\mathbf s^\mathbf{a}$ and frequency ratio $r_i$ (scalar). $\mathbf s^\mathbf{a}$ is passed through a 3D global average pooling (3D GAP) to collapse its spatial dimensions and then a 1D Conv to collapse its channel dimensions, so as to obtain a rough rPPG signal $\mathbf z \in \mathbb R^{1 \times T}$. 
 % \in \mathbb R^{T\times H \times W \times C}
%obtain the signal for color change over time. It is a mixture of the periodic rPPG signal and noises, 
%This is analogous to those hand-crafted signals of color change obtained in 
%obtained   Note that the 3D GAP layer has a similar goal with the handcrafted feature representations proposed in previous works 
%\cite{niu_rhythmnet_2020,ferrari_deepphys_2018}.
%, \ie to map the 4D video feature into 1D skin color variation signal. 
%The color change signal is passed through a 1D Conv to filter out noises and obtain a rough rPPG estimation 
%Meanwhile, the input scalar  $r_i$ is replicated by $T$ times to the same length of $\mathbf z_j$, denoted by 
%which has the same dimension of $1 \times T$ with $\mathbf z_j$. 
%$\mathbf r_i$. 
We aim to change the frequency of $\mathbf z$ from $f^a$ to $r_i \times f^a$. 
%It is served as a frequency multiplier vector that guides the frequency modulation for the local signal centered at every frame in $\mathbf z_j$. 
Directly multiplying $\mathbf z$ by $r_i$ only changes the signal amplitude but not the frequency.  
%represents the target frequency multiple that the signal $\mathbf z_j$ around the corresponding moment is supposed to be transformed into. 
%\st{while $r_i$ is temporally duplicated so that their outputs, \ie color variation signal $\mathbf z_j$ and temporal-wise target frequency feature $\mathbf r_i$, have the same dimension $1 \times T$.} 
%We replicate $r_i$ by $T$ times to the same length of $\mathbf z_j$ and concatenate them as the input to 
In practice, we replicate $r_i$ for $T$ times to create a vector $\mathbf r_i \in \mathbb R^{1 \times T}$
and then concatenate it with $\mathbf z$.
%\ZJ{Each value ($r_i$) of $\mathbf{r}_i$ indicates the frequency of corresponding local signal fragment of $\mathbf{z}$ should be modulated to $r_i\times f^a$.}
{Their concatenation} 
%\ZJ{The concatenation of $\mathbf{r}_i$ and $\mathbf{z}$} 
is processed via a 1D RB and a Bidirectional LSTM (BiLSTM). The Conv + ReLu in the 1D RB performs local transformation on the signal while the BiLSTM performs global transformation on it. This outputs a modulation vector $\mathbf m_{i} \in \mathbb R^{1 \times T}$, each of its entries can be seen as a multiplier to the corresponding component in $\mathbf z$, or equivalently corresponding frame in $\mathbf s^\mathbf{a}$. 
%In order to change the rPPG signal frequency of $\mathbf s^\mathbf{a}$ by a factor of $r_i$, 
We element-wisely multiply $\mathbf m_{i}$ to $\mathbf s^\mathbf{a}$ to change its rPPG signal frequency from $f^a$ to $r_i \times f^a$.
%, we apply  to it to obtain a $T$-dimensional vector, which can be seen as a coarse estimation of the rPPG signal inside $\mathbf s^\mathbf{a}_j$. For , we pass it through two 1D Conv to map to a vector of the same dimension $T$. 
%We aim to modulate the frequency of $\mathbf{y}^\mathbf{n}_i$ according to $r_i$. 
Note that, to obtain $\mathbf m_{i}$ from the concatenation of $\mathbf r_i$ and $\mathbf z$, the transformation needs to be non-linear (\ie 1D RB and BiLSTM). This is due to the nature of signal frequency modulation. We give an example: assuming we have 
%periodic signal $b^i$ can be represented as 
a periodic signal, $\alpha=A\sin (2\pi F (t+\varphi))$, where $A$, $F$ and $\varphi$ denote the signal amplitude, frequency and shift, respectively. To modulate its frequency by a factor of 2, \ie obtaining $\beta=A\sin (4\pi F (t+\varphi))$, we can compute the modulation vector as 
%\st{$A\sin (\pi F (t+\varphi))/A\sin (2\pi F (t+\varphi)) = 0.5\sec (\pi F (t+\varphi))$}
{$\beta/\alpha=2\cos (2\pi F (t+\varphi))$}, which indicates a non-linear {transformation}. The experimental analysis between non-linear transformation and linear transformation is presented in Sec.~\ref{sec5.1.3}.

\subsubsection{Video reconstruction loss}\label{sec3.2.2}
According to~\cite{yu_facial-video-based_2021}, the pixel values vary within a very small range across frames in the facial video. Hence, we need to prevent the significant color change even between $\mathbf{x^a}$ and its negative counterpart $\mathbf x^\mathbf{n}_i$. This will also help the subsequent signal extractors (\ie REA modules, Sec~\ref{sec3.3}) to distinguish the positive and negative samples by their insignificant color differences, hence extract their accurate underlying rPPG signals. To cope with this problem, we can regulate either the modulation weight $\|\mathbf m_{i} -1\|_2$ or the pixel-wise color distance between $\mathbf{x^a}$ and $\mathbf x^\mathbf{n}_i$ to be small. The latter works empirically better than the former in our experiment (Sec.~\ref{sec5.1.4}). We conceptualize the latter as the video reconstruction loss $ \mathcal L_\text{vr}$, which is indeed widely used in video frame inpainting \cite{szeto_temporally-aware_2020} and video super-resolution \cite{li_comisr_2021}, and write it out:
%and the anchor sample  {Given that the frequency of human rPPG signal lies in a small range from 0.75 Hz to 3 Hz \cite{yue_multimodal_2022}, we attempt to avoid excessive color differences between the generated negative samples $\mathbf{X^n}=\{\mathbf{x}^n\}$ and the anchor sample $\mathbf{x^a}$. 
%To achieve this goal, we have two solutions. The first one is to constrain each entry of the modulation vector $\mathbf m_{i}$ to be as closer to $1$ as possible, thus the FMB block will not produce too much color differences between its output and its input $\mathbf s^\mathbf{a}$. The second one is to constrain the pixel-wise distance between $\mathbf{x}^\mathbf{n}_i$ and $\mathbf{x^a}$. 
%We empirically take the second one as our solution and report the performance of the first one in Sec.~\ref{sec4.4.4}. Specifically, we employ the video reconstruction loss, which is widely used in video frame inpainting \cite{szeto_temporally-aware_2020} and video super-resolution \cite{li_comisr_2021} , for the reinforcement of our LEA module. 
\begin{equation}\label{equ1}
    \mathcal L_\text{vr}= {\textstyle \frac{1}{k} \sum_{i=1}^{k}}\left \| \mathbf{x}^\mathbf{n}_i-\mathbf{x^a} \right \|_2
\end{equation}
{This loss enforces the LFA module to augment negative samples which retain the general visual appearance 
%\ZJ{and DC component} 
of $\mathbf{x^a}$.}

\subsection{Signal extraction: local rPPG expert aggregation}\label{sec3.3}

\begin{figure*}[!t]
  \centerline{\includegraphics[width=7.2in]{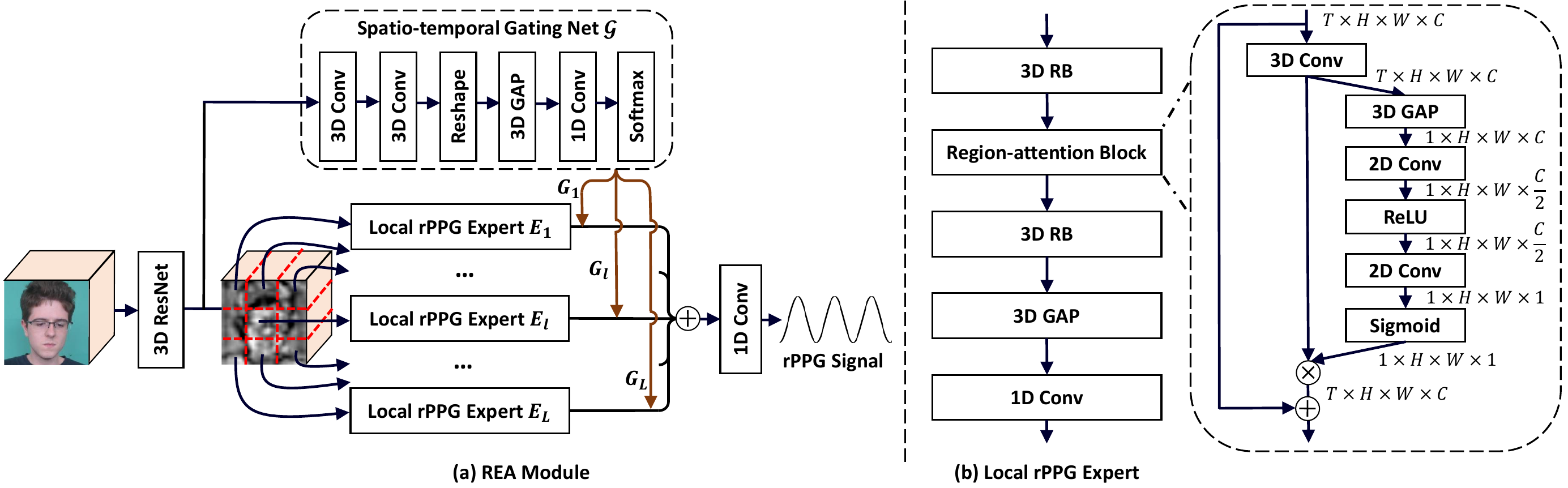}}
  %\vspace{-0.1in}
  \caption{(a) The local rPPG expert aggregation (REA) module captures complementary pulsation information from different face regions. We regard the rPPG estimation from each local region as a local rPPG expert. Different experts are aggregated through a spatio-temporal gating net. (b) The structure for obtaining one local rPPG expert.}
  \label{fig4}
  \vspace{-0.1in}
  \end{figure*}

% \begin{figure}[!t]
%   \centerline{\includegraphics[width=2.8in]{figs/fig_major1.pdf}}
%   %\vspace{-0.1in}
%   \caption{\ZJ{The structure of the proposed spatio-temporal gating net $\mathcal{G}$.}}
%   \label{fig13}
%   \vspace{-0.1in}
%   \end{figure}
  
% \begin{figure}[!t]
%   \centerline{\includegraphics[width=2.7in]{figs/fig5.png}}
%   %\vspace{-0.1in}
%   \caption{The structure of region-attention block. It generates spatial attention map to enhance physiological information from pulsation-sensitive regions and suppress information from pulsation-insensitive regions.}
%   \label{fig5}
%   \vspace{-0.1in}
%   \end{figure}

% We design the Spatial-attention mixture of experts (SA-MoE) as an rPPG estimator aiming to estimate rPPG signals from given samples. 
We extract the rPPG signal from each augmented sample via the local rPPG expert aggregation (REA) module. 
Our motivation is that different face regions have varying distributions of {blood vessels} and noises and should contribute differently to the rPPG estimation. Our REA module is designed to combine complementary pulsation information from these regions for accurate rPPG estimation.

%The rPPG estimator aims to estimate rPPG signals from given samples. 
The structure of the module is illustrated in Fig.~\ref{fig4}a. We first leverage the 3D ResNet-10 \cite{yue_deep_2021} to encode the input video into a feature tensor. 
%Then, \st{considering that different skin regions have different contribution levels to the estimation result because of their varying density degrees of blood vessels and different noise distributions} 
% \cite{lu_dual-gan_2021}
The feature tensor is then divided into $L$ evenly partitioned regions where we extract rPPG signal from each of them and denote it as a local rPPG expert $E_l$. As illustrated in  Fig.~\ref{fig4}b,   
%$D_l\in \mathbb{R}^{T\times H\times W\times C},l\in \left \{1,2,\dots,L \right \}$ (REA module with $L=4$ is shown in Fig.~\ref{fig4}(a) as an exemplar). $H,W,C$ respectively indicates the height, width and channel of the divided local feature. 
%Each region 
% has its individual vascular and noise distributions thus 
%is performed with a region-attention expert (RA-Expert) $E_l$ to handle its noises and extract underlying rPPG signal. 
$E_l$ is obtained via two 3D RBs for feature processing; one region-attention block in-between the two 3D RBs for capturing physiological clues from pulsation-sensitive skins; and one 3D GAP followed by one 1D Conv for projecting the feature into 1D signal. Having $E_1, E_2,\dots, E_L$ for $L$ regions, 
%After estimating signals from $L$ regions via experts $E_1,E_2,\dots,E_L$ accordingly, 
we design a spatio-temporal gating net $\mathcal{G}$ to aggregate different experts into one rPPG prediction. 

Note that although the spatial augmentation we adopt for positive sample generation can change the spatial layout of face regions, we re-arrange the experts according to the same spatial augmentation to maintain the correspondence between the face regions and local rPPG experts. During testing, spatial augmentation is no longer used, we can directly connect face regions to their corresponding local rPPG experts.

%For example, for the frontal face in the negative sample (without spatial augmentation, an example is shown in Fig.~\ref{fig4}a), the nose region is processed by the $l$-th expert model. For the positive samples (with image flips or rotations), we also ensure that their nose regions are processed by the $l$-th model.}
Below we specify our region-attention block and spatio-temporal gating net.

\subsubsection{Region-attention block} Within each face region, some pixels may belong to the background that do not contribute to rPPG estimation; while some other pixels, despite on the face, are insensitive to blood pulsation (\eg {eyes and mouth}). %We specify the motivation of RA-Block is that each pixel in $D_l$ contributes at different levels for rPPG estimation ().
Region-attention block is thereby introduced to focus rPPG estimation on the pulsation-sensitive skin area. Fig.~\ref{fig4}b illustrates its structure: given the 4D input feature tensor {($\mathbb R^{T \times H \times W \times C}$)}, after a 3D Conv, we leverage a 3D GAP to collapse it along the temporal dimension into a 3D tensor {($\mathbb R^{1 \times H \times W \times C}$)}. Inspired by the squeeze and excitation operation in \cite{hu_squeeze-and-excitation_2020}, we pass the feature into consecutive 2D Conv and non-linear activation layers (\ie ReLu, Sigmoid) for aggregating information across the channel dimension and obtaining the region-attention map. The attention map is pixel-wisely multiplied to the input 4D feature. A skip connection is also leveraged to retain the input information. Our scheme indeed transforms the image-based channel-wise attention in \cite{hu_squeeze-and-excitation_2020} into video-based spatial-wise attention, which enhances the pulsation-sensitive area and {suppresses} background and pulsation-insensitive area in rPPG estimation. The superiority of the proposed region-attention block is experimentally demonstrated in Sec.~\ref{sec5.2.1}.
%over another skin-segmentation based attention mechanism 

\subsubsection{Spatio-temporal gating net} \label{sec:gatingnet}
The idea of employing a gating net to combine experts is from mixture of experts \cite{jacobs_adaptive_1991}. The conventional design of gating net \cite{emad_moesr_2022,sahasrabudhe_deep_2021} normally assigns sample-level weights among experts. Unlike them, we introduce the spatio-temporal gating net $\mathcal{G}$ which assigns spatio-temporal weights to combine temporal rPPG signals (experts) extracted from different spatial regions. Our motivation is that some regions may reflect rPPG signal better during the systole while some are more sensitive during the diastole. Each expert signal should be assigned with different weights at different moments.  
%This is indeed the designing spirit of our spatio-temporal gating net $\mathcal{G}$. 
The structure of $\mathcal G$ is illustrated in Fig.~\ref{fig4}a, consisting of two 3D Convs followed by a Reshape, a 3D GAP, a 1D Conv and a Softmax layer, respectively. 
% \ZJ{There is a Reshape layer integrated in $\mathcal G$ to divide the feature map into $L$ regions corresponding to $L$ experts. The gating net respectively analyses the sensitivities of partitioned regions for reflecting physiological clues. Then it } 
The gating net generates $L$ vectors, $G_1=\left \{ g_1^1,\dots,g_1^T \right \}$, $\dots$, $G_L=\left \{ g_L^1,\dots,g_L^T \right \}$, corresponding to $L$ experts $E_1,\dots, E_L$. Each vector is of dimension {$1 \times T$}, equivalent to the temporal dimension of each expert. Vector components of the same index $t$ are softmaxed across experts, \ie $softmax\left (g_1^t,\dots,g_L^t \right )$, to ensure that, at the $t$-th moment, the sum of weights assigned to all experts to be 1. We element-wisely multiply each $G_l$ to its corresponding $E_l$ and obtain the final rPPG signal as a weighted combination: 
\begin{equation}\label{equ2}
    \mathbf y= {\textstyle \sum_{l=1}^{L} E_l \cdot G_l}
\end{equation}
% the estimated $L$ signals are weighted summed up to obtain the final signal: $y= {\textstyle \sum_{l=1}^{L} E_l(D_l)\times G_l}$.

%\ZJ{The main structure of REA is  from the mixture of experts (MoE) \cite{jacobs_adaptive_1991}.} The original MoE deploys a gating net $\mathcal{G}$ to assign different weights $G_1,G_2,\dots,G_L$ to multiple experts $E_1,E_2,\dots,E_L$ for controlling their activations. The output $E_{out}$ is the weighted sum of these experts for information fusion: $E_{out}={\textstyle \sum_{l=1}^{L}E_{l} \times G_{l}}$. 

The REA module is devised for positive and negative video samples to produce positive signals $\mathbf{Y^p}=\left \{\mathbf{y}^\mathbf{p}_1,\mathbf{y}^\mathbf{p}_2 \right\}$ and negative signals $\mathbf{Y^n}=\left \{\mathbf{y}^\mathbf{n}_1,\mathbf{y}^\mathbf{n}_2,\dots,\mathbf{y}^\mathbf{n}_k\right \}$. These signals are of various frequencies and can be used for subsequent self-supervised training.

\subsection{Network optimization: frequency-inspired losses}\label{sec3.4}

We introduce a series of frequency-inspired losses, \ie frequency contrastive loss, frequency ratio consistency loss, cross-video frequency agreement loss, for the
optimization of rPPG signals generated from
the input video's augmentations and neighbors.  

%We leverage four loss functions to optimize the network parameters: the reconstruction loss, the frequency contrastive loss, the frequency ratio consistency loss and the cross-video frequency agreement loss. 
%\ZJ{The latter three frequency-inspired losses aims to optimize multiple rPPG signals generated from the input video, its augmentations, and its neighbors.}

\subsubsection{Frequency contrastive loss}\label{sec3.4.1}
Since we do not have ground truth PPG signals for training,  
%\ZJ{aim to train the model without using signal ground truth}, 
we use the popular contrastive self-supervised learning~\cite{oord_representation_2019,he_momentum_2020} to pull close rPPG signals ($\mathbf {Y^p}=\{\mathbf{y^p}\}$) from positive samples while pushing them away from rPPG signals ($\mathbf {Y^n}=\{\mathbf{y^n}\}$) from negative samples in the feature space. We adapt the InfoNCE loss~\cite{oord_representation_2019} to write out our frequency contrastive loss $\mathcal L_\text{fc}$:  
%equency contrastive loss $l_{con}$ for self-supervised training:
\begin{equation}\label{equ3}
\mathcal L_\text{fc}=\log(\frac{\exp(d(\mathbf{y}^\mathbf{p}_1,\mathbf{y}^\mathbf{p}_2)/\tau )}{ {\textstyle \sum_{i=1}^{k}} (\exp(d(\mathbf{y}^\mathbf{p}_1,\mathbf{y}^\mathbf{n}_i)/\tau )+\exp(d(\mathbf{y}^\mathbf{p}_2,\mathbf{y}^\mathbf{n}_i)/\tau ))}+1)
\end{equation}
where $d(\cdot,\cdot)$ measures the frequency difference between two signals and $\tau$ is the temperature. We follow~\cite{gideon_way_2021} to realize $d$ as the mean squared error between power spectral densities (PSD) of two signals. PSD describes the power of different frequency components in a signal. Notice \cite{gideon_way_2021} utilizes the triplet loss to optimize among only three {signals}, \ie anchor, positive and negative; we have a different loss form as a result of our larger numbers of positive and negative {signals} generated by different ways.
%$l_{fcl}$ serves as an self-supervised objective function to pull positive signals $Y_p=\left \{y_p^1,y_p^2 \right\}$ together, while separating them with negative signals $Y_n=\left \{y_n^1,y_n^2,\dots,y_n^k\right \}$. 
%Therefore, $l_{fcl}$ can enforce our rPPG estimator to learn to detect whether the input videos contain similar or dissimilar signals, and train it to capture the most important frequency information of input videos for rPPG estimation.

\subsubsection{Frequency ratio consistency loss}\label{sec3.4.2}
This loss constrains that the frequency ratio between any positive and negative rPPG signals ($\mathbf{y^p}$ and $\mathbf{y^n}$) should be consistent with the corresponding ratio $r$ fed into the LFA module. It optimizes the LFA module to generate a certain negative sample with the target signal frequency.
We define the frequency ratio consistency loss $\mathcal L_\text{fr}$:
\begin{equation}\label{equ4}
\mathcal L_\text{fr}= {\textstyle \frac{1}{2k} \sum_{i=1}^{k}\left |\frac{P(\mathbf{y}^\mathbf{n}_i)}{P(\mathbf{y}^\mathbf{p}_1)}-r_i\right |+\left |\frac{P(\mathbf{y}^\mathbf{n}_i)}{P(\mathbf{y}^\mathbf{p}_2)}-r_i\right | } 
\end{equation}
where we use $P(\cdot)$ to measure a signal's dominant frequency. $P$ is implemented by following \cite{poh_non-contact_2010} to apply the fast Fourier transform on a given signal and select the frequency component with maximum power. $\mathbf{y}^\mathbf{p}_1$ and $\mathbf{y}^\mathbf{p}_2$ should ideally have the same frequency. $\frac{P(\mathbf{y}^\mathbf{n}_i)}{P(\mathbf{y}^\mathbf{p}_1)}$ indicates the dominant frequency ratio between signal $\mathbf{y}^\mathbf{n}_i$ and $\mathbf{y}^\mathbf{p}_1$. If this ratio is equal to $r_i$, that means the LFA module has successfully generated $\mathbf{x}^\mathbf{n}_i$ with the target {rPPG signal} frequency $r_i \times f^a$.
%By minimizing $\mathcal L_\text{fr}$, 
% performs frequency transformation and
% By constrains the frequency ratios between $Y_n$ and $Y_p$, 
%we can guide LFA to correctly transform the signal to its target high/low frequency version, thus achieving frequency augmentation.

\subsubsection{Cross-video frequency agreement loss}\label{sec3.4.3}
This loss function is built upon the observation that the frequency of an rPPG signal does not change rapidly over a short term. Based on this, the rPPG signal from the input $\mathbf{x^a}$ should have a similar frequency to those from $\mathbf{x^a}$'s temporal neighbors (see Sec.~\ref{sec3.1}). We pass $\mathbf{x^a}$'s neighboring video clips through REA modules to extract rPPG signals from them, \ie $\mathbf{Y^b} = \{\mathbf{y^b}\}$. We design the cross-video frequency agreement loss $\mathcal L_\text{fa}$ to enforce the frequency conformability between $\mathbf{Y^p}$ and $\mathbf{Y^b}$:
\begin{equation}\label{equ5}
\mathcal L_\text{fa}= {\textstyle \frac{1}{2J} \sum_{j=1}^{J}d(\mathbf{y}^\mathbf{p}_1,\mathbf{y}^\mathbf{b}_j)+d(\mathbf{y}^\mathbf{p}_2,\mathbf{y}^\mathbf{b}_j)} 
\end{equation}
%where $y_b$ represents the signal estimated from the $b$-th clip of the face video and 
%there are $B$ non-overlapping clips in total. 
where $d(\cdot,\cdot)$ is the same to that in Eqn.~\ref{equ3}. We have $J$ neighbors in total. 
This loss helps reduce the signal estimation error and improve the signal periodicity.
%This loss helps enhance the periodicity of the estimated signal. 

\medskip 
The overall loss function is a linear combination of them as well as the video reconstruction loss in Sec.~\ref{sec3.2.2}:
\begin{equation}\label{equ6}
\mathcal L = \mathcal L_\text{fc}+\mathcal L_\text{fr}+\mathcal L_\text{fa}+\mathcal L_\text{vr}
\end{equation}
The effectiveness of these frequency-inspired losses is evaluated in Sec.~\ref{sec5.3}.

\section{Experiments}

\subsection{Datasets}\label{sec4.1}

We conduct our experiments on five public datasets: UBFC-rPPG \cite{bobbia_unsupervised_2019}, PURE \cite{stricker_non-contact_2014}, DEAP \cite{koelstra_deap_2012},  MMVS \cite{yue_deep_2021}, and BP4D+ \cite{zhang_multimodal_2016}.

\para{UBFC-rPPG} consists of 42 facial videos with simultaneously recorded PPG signals and heart rates. The resolution and frame rate of each facial video is $640\times480$ and 30 frames per second (FPS), respectively. We follow \cite{lee_meta-rppg_2020} to discard subjects of indices 11, 18, 20 and 24 because their heart rates were inappropriately recorded. 

\para{PURE} contains 60 facial videos from 10 subjects. During the data collection process, subjects were asked to perform six kinds of head motions (small rotation, medium rotation, slow translation, fast translation, talking and steady) in front of the camera for one minute. The videos were captured at a frame rate of 30 FPS and a resolution of $640\times480$. The ground truth PPG signals were recorded using a finger clip pulse oximeter with a sampling rate of 60 Hz. We follow \cite{gideon_way_2021} to discard the first two samples because their PPG waveforms were strongly corrupted.

\para{DEAP} consists of 874 facial videos associated with multi-channel physiological signals. They were taken from 22 subjects by playing one-minute musical excerpts to them. Each facial video has a resolution of $720\times576$ and a frame rate of 50 FPS. 

\para{MMVS} contains 745 facial videos from 129 subjects. A depth camera and a finger clip pulse oximeter was utilized to record facial videos and PPG signals, respectively. The resolution of each captured video is $1920\times1080$ and the frame rate is 25 FPS. The sampling rate for PPG signals is 60 Hz.

\para{BP4D+} contains 548 facial videos from 140 subjects. There are 58 male subjects and 82 female subjects, with their ages ranging from 18 to 66. Ethnicities of these subjects include Caucasian, African and Asian (East-Asian and Middle-East-Asian); specifically, 64 Caucasian, 46 Asian, and 15 African subjects. The resolution and frame rate of each video is $1092\times1340$ and 25 FPS, respectively.
%We leverage this dataset to evaluate the performance of our approach on subjects of different ethnicities.}

%More dataset details can be found in \cite{yue_deep_2021}.

\subsection{Implementation Details}\label{sec4.2}
{For} facial videos in five datasets, we randomly sample consecutive 600 frames from each video and follow \cite{gideon_way_2021} to scale them into a resolution of $64\times64$ for training. We cut the 600-frame video into four clips, each with an equivalent length $T$ of 150 frames. In every iteration, we randomly select one clip as the main input $\mathbf{x^a}$ while the rest three  ($J = 3$ in Eqn.~\ref{equ5}) 
are its neighbors.
% for} \ZJ{The default number of video clips ($J+1$, $J=3$ in Eqn.~\ref{equ5}) is equivalent to that of \cite{yu_physformer_2022}, and the default length of video clips $T=150$ is equivalent to that of \cite{song_heart_2020}.
The number of negative samples, $k$, is set to 4. The number of face regions, $L$, is set to 9. The temperature $\tau$ in Eqn.~\ref{equ3} 
is set to 0.08. Most parameter settings in this paper follow the common practices in the literature. For instance, $k$ is equivalent to that in~\cite{kim_self-supervised_2021}; $L$ is inspired by \cite{yu_physformer_2022}; $\tau$ is the same to that in \cite{diba_vi2clr_2021}; while $J$ and $T$ are equivalent to that in \cite{yu_physformer_2022} and \cite{song_heart_2020}, respectively.  

% The loss weights $\lambda_1, \lambda _2, \lambda _3, \lambda _4$ are 0.5, 1, 0.5, 0.5 respectively. 
Our model is trained for 100 epochs on four NVIDIA GeForce RTX 2080 GPUs using Pytorch 1.8.0. We use the Adam \cite{kingma_adam_2017} optimizer and set the batch size to 4. The learning rate is initialized as $1 \times 10^{-5}$ and is decreased to $0.5 \times 10^{-5}$ at the 50-th epoch.

\subsection{Evaluation Protocol}\label{sec4.3}

{Previous works calculate the heart rate (HR), heart rate variability (HRV) and respiration frequency (RF) from estimated rPPG signals and compare them to the corresponding ground truth for performance evaluation \cite{lu_dual-gan_2021,song_pulsegan_2021,niu_video-based_2020,yu_remote_2019}. We follow them to conduct HR evaluation on five datasets; HRV and RF evaluation on the UBFC-rPPG dataset. Moreover, we perform cross-dataset HR evaluation among UBFC-rPPG, PURE and MMVS. The calculation of HR, HRV and RF is via the toolkit HeartPy \cite{van_gent_analysing_2019}.}

We follow \cite{song_pulsegan_2021,gideon_way_2021} to use mean absolute error (MAE), root mean square error (RMSE) and Pearson's correlation coefficient ($r$) as evaluation metrics for HR. For HRV, we follow \cite{lu_dual-gan_2021,yu_physformer_2022,niu_video-based_2020} to compute its three attributes, \ie low frequency (LF), high frequency (HF), and LF/HF ratio. LF and HF are calculated from the interbeat intervals under low frequency (0.04 to 0.15 Hz) and high frequency (0.15 to 0.4 Hz) bands of the rPPG signal \cite{yu_physformer_2022,lu_dual-gan_2021}. 
%They have normalized units (n.u.). 
For each attribute of HRV, we report the standard deviation (Std) of estimation errors, RMSE and $r$. Finally, for RF, we also report the Std, RMSE and $r$ as per most comparable methods \cite{lu_dual-gan_2021,niu_video-based_2020,yu_remote_2019}. We follow \cite{yue_deep_2021,lu_dual-gan_2021,niu_video-based_2020} to perform the 5-fold subject-exclusive cross-validation for all experiments. 

\begin{table*}[!t]
\caption{Comparison to state of the art on HR estimation. The results are reported on UBFC-rPPG, PURE, DEAP, MMVS and BP4D+ datasets. $\uparrow$ indicates that the larger the value is the better it is and $\downarrow$ vice versa. The best supervised approach is marked in \colorbox{gray!50}{shadow}, while the best self-supervised approach is marked in \textbf{bold}.}
\vspace{-0.1in}
\label{table1}
\begin{center}
\setlength{\tabcolsep}{0.7mm}
\begin{tabular}{c|c|ccc|ccc|ccc|ccc|ccc}
\toprule
\multirow{2}*{Method} &PPG &\multicolumn{3}{c|}{UBFC-rPPG}& \multicolumn{3}{c|}{PURE}& \multicolumn{3}{c|}{DEAP}& \multicolumn{3}{c|}{MMVS} & \multicolumn{3}{c}{BP4D+}\\ 
& annotations&MAE$\downarrow$  &RMSE$\downarrow$  &$r\uparrow$  & MAE$\downarrow$  &RMSE$\downarrow$  &$r\uparrow$ & MAE$\downarrow$  &RMSE$\downarrow$  &$r\uparrow$ & MAE$\downarrow$  &RMSE$\downarrow$  &$r\uparrow$ & MAE$\downarrow$  &RMSE$\downarrow$  &$r\uparrow$\\ 
\midrule
POS \cite{wang_algorithmic_2017} &- &8.35  &10.00  &0.24  &3.14  &10.57  &0.95  &7.39  &10.25  &0.82 &6.77 &9.40 &0.82 &6.82 &9.70 &0.77\\
CHROM \cite{de_haan_robust_2013} &- &8.20  &9.92  &0.27  &3.82  &6.80  &0.97  &7.47  &10.31  &0.82 &6.85 &9.37  &0.82 &7.25 &9.91 &0.72\\
Green \cite{verkruysse_remote_2008} &- &6.01  &7.87  &0.29  &4.39  &11.60  &0.90  &8.10  &11.17  &0.80 &7.13 &9.46  &0.80 &6.76 &9.73 &0.76\\
SynRhythm \cite{niu_synrhythm_2018} &$\surd$  &5.59  &6.82  &0.72  &2.71  &4.86 &0.98  &5.08  &5.92  &0.87 &4.48 &6.52  &0.89 &4.33 &6.15 &0.85\\
Meta-rppg \cite{lee_meta-rppg_2020} &$\surd$ &5.97  &7.42  &0.53  &2.52  &4.63 &0.98  &5.16  &6.00  &0.87 &4.30 &6.20  &0.91 &4.08 &6.11 &0.86\\
PulseGan \cite{song_pulsegan_2021} &$\surd$ &1.19 &2.10  &0.98  &2.28  &4.29 &0.99  &4.86 &5.70  &0.88 &3.52 &5.09  &0.93 &3.84 &5.37 &0.89\\ 
Dual-Gan \cite{lu_dual-gan_2021} &$\surd$ &0.44  &\colorbox{gray!50}{0.67}  &0.99  &\colorbox{gray!50}{0.82} &\colorbox{gray!50}{1.31}  &\colorbox{gray!50}{0.99} &3.25  &4.11  &0.91 &\colorbox{gray!50}{3.00} &\colorbox{gray!50}{4.27} &\colorbox{gray!50}{0.94} &\colorbox{gray!50}{2.96} &\colorbox{gray!50}{4.18} &\colorbox{gray!50}{0.93}\\
Physformer \cite{yu_physformer_2022} &$\surd$ &\colorbox{gray!50}{0.40} &0.71  &\colorbox{gray!50}{0.99}  &1.10  &1.75 &0.99  &\colorbox{gray!50}{3.03} &\colorbox{gray!50}{3.96}  &\colorbox{gray!50}{0.92} &3.28 &4.50  &0.93 &3.10 &4.33 &0.92\\ 
\midrule
Gideon \etal \cite{gideon_way_2021} &$\times$ &1.85  &4.28  &0.93  &2.32  &2.97  &0.99  &5.13  &6.16  &0.86 &3.43 &4.74 &0.93 &4.09 &5.60 &0.85\\
Ours &$\times$ &\textbf{0.58}  &\textbf{0.94} &\textbf{0.99} &\textbf{1.23} &\textbf{2.01}  &\textbf{0.99}  &\textbf{4.20}  &\textbf{5.18} &\textbf{0.90} &\textbf{2.93} &\textbf{4.16}  &\textbf{0.94} &\textbf{3.22} &\textbf{4.47}  &\textbf{0.91}\\ 

\bottomrule
\end{tabular}
\end{center}
\end{table*}

\subsection{Results}
\subsubsection{HR evaluation}

We first evaluate the HR estimation on five datasets. Representative approaches include: 1) traditional blind signal separation- and skin reflection model-based ones: POS \cite{wang_algorithmic_2017}, CHROM \cite{de_haan_robust_2013} and Green \cite{verkruysse_remote_2008}; 2) modern DNN-based supervised ones: SynRhythm \cite{niu_synrhythm_2018}, Meta-rppg \cite{lee_meta-rppg_2020}, PulseGan \cite{song_pulsegan_2021}, Dual-Gan \cite{lu_dual-gan_2021}, Physformer \cite{yu_physformer_2022}; 
and 3) a recent DNN-based self-supervised one: Gideon \etal \cite{gideon_way_2021}. Their results are shown in Table \ref{table1}. 

First, we observe that the performance of traditional approaches \cite{wang_algorithmic_2017,de_haan_robust_2013,verkruysse_remote_2008} is much inferior to DNN-based ones. DNN-based approaches can be trained in a data-driven way, while traditional approaches heavily rely on priors, which can not always satisfy. Second, DNN-based self-supervised approaches (\cite{gideon_way_2021} and ours) show comparable performance to many supervised ones, \eg \cite{niu_synrhythm_2018,lee_meta-rppg_2020,song_pulsegan_2021,lu_dual-gan_2021,yu_physformer_2022}. 
%, especially on the PURE and MMVS datasets. 

Our approach outperforms many supervised approaches. 
For instance, on the MMVS dataset, it even decreases the MAE/RMSE from best performing one Dual-Gan \cite{lu_dual-gan_2021} by 0.07/0.11. 
MMVS is a large-scale dataset in which videos were captured from different scenarios. 
%; while our self-supervised approach can be good at extracting domain-generic information (see also Sec.~\ref{tab:cross-dataset}); 2) 
We reproduce the state of the art on MMVS using their default parameters, which may need to be adjusted on this large-scale dataset. In contrast, our self-supervised model has demonstrated good generalizability on this dataset.  
%Our model, despite being trained without PPG annotations, has successfully learned to capture accurate rPPG signals from face videos. 

Our model also significantly outperforms the very recent self-supervised approach Gideon \etal \cite{gideon_way_2021} on all five datasets; for example, on UBFC-rPPG, it decreases MAE by 1.27 and RMSE by 3.34. This justifies the effectiveness of our proposed modules and losses. 

\begin{figure}[!t]
  \centerline{\includegraphics[width=3.5in]{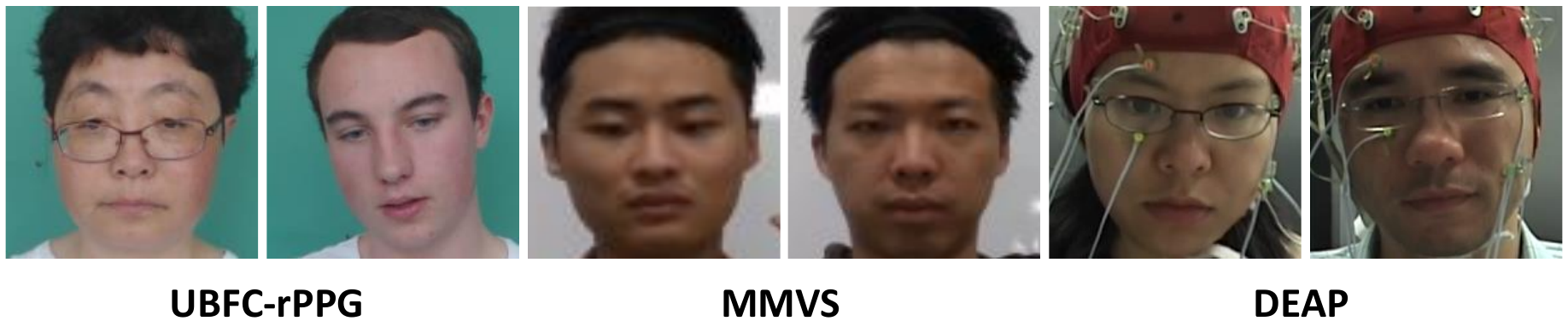}}
  %\vspace{-0.1in}
  \caption{Face images in UBFC-rPPG, MMVS, and DEAP datasets.}
  \label{fig18}
 % \vspace{-0.1in}
  \end{figure}

It is worth noting that the general performance of comparable methods on the DEAP dataset is clearly inferior to that on the rest datasets. The reason is that, during the data collection process of the DEAP dataset, subjects wore many devices on their heads to record multi-channel physiological signals, such as EEG, EOG and EMG signals. Devices like EEG caps, electrodes and wires on the heads cover many skin areas, thus hindering the capturing of physiological clues for rPPG estimation. While subjects in other datasets did not wear these devices. Some face images in UBFC-rPPG, MMVS and DEAP datasets are shown in Fig.~\ref{fig18}.

%module to produce more effective negative samples with 
%rich frequency diversity.} 
%\st{improvement of generating positive signals from positive samples instead of negative signals to maintain their frequency consistency.} 
%\ZJ{Also, these results indicate the the advantage of our local rPPG expert aggregation module (REA) to} \st{multiple frequency-modulated negative samples for frequency contrastive learning can promote the estimator to encode more discriminative features; 2) learning and integrating} \ZJ{capture complementary pulsation clues from different face regions for better estimation performance.} \st{best the estimation performance; 3) the proposed SA-block, frequency ratio loss $l_{frl}$, projection loss $l_{pro}$ can improve the representation learning ability of FISL (ablation study in section IV.D.4)).}

\begin{table*}[!t]
\caption{Comparison to state of the art on RF and HRV estimation.  The results are reported on the UBFC-rPPG. $\uparrow$ indicates that the larger the value is the better {it is} and $\downarrow$ vice versa. The best supervised approach is marked in \colorbox{gray!50}{shadow}, while the best self-supervised approach is  marked in \textbf{bold}.}
\vspace{-0.1in}
\label{table2}
\begin{center}
\setlength{\tabcolsep}{1mm}
\begin{tabular}{c|c|ccc|ccc|ccc|ccc}
\toprule
% &&&&&\multicolumn{9}{c}{} \\ 
\multirow{2}*{\diagbox{Method}{UBFC-rPPG}} &PPG&\multicolumn{3}{c|}{RF}& \multicolumn{3}{c|}{HRV: LF}& \multicolumn{3}{c|}{HRV: HF}& \multicolumn{3}{c}{HRV: LF/HF}\\

& annotations&Std$\downarrow$  &RMSE$\downarrow$  &$r\uparrow$  & Std$\downarrow$  &RMSE$\downarrow$  &$r\uparrow$ & Std$\downarrow$  &RMSE$\downarrow$  &$r\uparrow$ & Std$\downarrow$  &RMSE$\downarrow$  &$r\uparrow$\\ 
\midrule
POS \cite{wang_algorithmic_2017} &- &0.109  &0.107  &0.087  &0.171  &0.169  &0.479  &0.171  &0.169  &0.479 &0.405 &0.399 &0.518\\
CHROM \cite{de_haan_robust_2013} &-&0.086  &0.089  &0.102  &0.243  &0.240  &0.159 &0.243  &0.240  &0.159 &0.655 &0.645  &0.266\\
Green \cite{verkruysse_remote_2008} &-&0.087  &0.086  &0.111  &0.186  &0.186  &0.280  &0.186  &0.186  &0.280 &0.361 &0.365  &0.492\\
CVD  \cite{niu_video-based_2020} &$\surd$ &0.017  &0.018  &0.252  &0.053  &0.065 &0.740  &0.053  &0.065  &0.740 &0.169 &0.168  &0.812\\
rPPGNet \cite{yu_remote_2019} &$\surd$ &0.030  &0.034  &0.233  &0.071 &0.070 &0.686  &0.071  &0.070 &0.686 &0.212 &0.208  &0.744\\
Dual-Gan \cite{lu_dual-gan_2021} &$\surd$ &0.010  &0.010  &0.395  &0.034 &0.035  &0.891  &0.034  &0.035  &0.891 &0.131 &0.136  &0.881\\
Physformer \cite{yu_physformer_2022} &$\surd$ &\colorbox{gray!50}{0.009}  &\colorbox{gray!50}{0.009}  &\colorbox{gray!50}{0.413}  &\colorbox{gray!50}{0.030} &\colorbox{gray!50}{0.032} &\colorbox{gray!50}{0.895}  &\colorbox{gray!50}{0.030} &\colorbox{gray!50}{0.032} &\colorbox{gray!50}{0.895} &\colorbox{gray!50}{0.126} &\colorbox{gray!50}{0.130}  &\colorbox{gray!50}{0.893}\\
\midrule
Gideon \etal \cite{gideon_way_2021} &$\times$ &0.061  &0.098  &0.103  &0.091  &0.139  &0.694  &0.091  &0.139  &0.694  &0.525 &0.691  &0.684\\
Ours &$\times$ &\textbf{0.023}  &\textbf{0.028}  &\textbf{0.351}  &\textbf{0.047}  &\textbf{0.062}  &\textbf{0.769}  &\textbf{0.047}  &\textbf{0.062}  &\textbf{0.769} &\textbf{0.160} &\textbf{0.164}  &\textbf{0.831}\\ 

\bottomrule
\end{tabular}
\end{center}
\end{table*}

\begin{table*}[!t]
\caption{Comparison to state of the art on cross-dataset HR estimation.  $\uparrow$ indicates that the larger the value is the better {it is} and $\downarrow$ vice versa. The best supervised approach is marked in \colorbox{gray!50}{shadow}, while the best self-supervised approach is  marked in \textbf{bold}.}
\vspace{-0.1in}
\label{table4}
\begin{center}
\setlength{\tabcolsep}{1.5mm}
\begin{tabular}{c|c|ccc|ccc|ccc|ccc}
\toprule
\multirow{2}*{Method}  &\multirow{2}*{PPG annotations} &\multicolumn{3}{c|}{MMVS$\rightarrow$UBFC-rPPG} &\multicolumn{3}{c|}{UBFC-rPPG$\rightarrow$MMVS}&\multicolumn{3}{c|}{PURE$\rightarrow$UBFC-rPPG}&\multicolumn{3}{c}{UBFC-rPPG$\rightarrow$PURE}\\ 
&&MAE$\downarrow$  &RMSE$\downarrow$  &$r\uparrow$  & MAE$\downarrow$  &RMSE$\downarrow$  &$r\uparrow$ & MAE$\downarrow$  &RMSE$\downarrow$  &$r\uparrow$ & MAE$\downarrow$  &RMSE$\downarrow$  &$r\uparrow$\\ 
\midrule
Meta-rppg  \cite{lee_meta-rppg_2020} &$\surd$&6.48 &7.97 &0.52 &5.69  &7.74  &0.84 &6.11 &7.58 &0.66 &4.00 &5.98 &0.92\\
PulseGan  \cite{song_pulsegan_2021} &$\surd$&2.33 &3.62 &0.97 &4.40  &6.35  &0.89  &2.30 &3.50 &0.97 &3.36 &5.11 &0.95 \\
Dual-Gan \cite{lu_dual-gan_2021} &$\surd$&2.00 &3.13 &0.97 &3.51  &4.99  &0.93 &2.03 &\colorbox{gray!50}{3.01} &0.97 &\colorbox{gray!50}{1.81} &\colorbox{gray!50}{2.97} &\colorbox{gray!50}{0.99} \\
% RhythmNet \cite{niu_rhythmnet_2020} &$\surd$&4.02 &5.35 &0.86 &4.82  &6.63  &0.87 &3.65 &4.44 &0.90 &3.72 &5.75 &0.95 \\
Physformer \cite{yu_physformer_2022} &$\surd$&\colorbox{gray!50}{1.97} &\colorbox{gray!50}{3.08} &\colorbox{gray!50}{0.97} &\colorbox{gray!50}{3.46}  &\colorbox{gray!50}{4.96}  &\colorbox{gray!50}{0.93} &\colorbox{gray!50}{1.93} &3.02 &\colorbox{gray!50}{0.97} &1.99 &3.28 &0.99 \\
\midrule
Gideon \etal \cite{gideon_way_2021} &$\times$&2.45 &3.71 &0.93 &3.84  &5.31  &0.91 &2.37 &3.51 &0.95 &2.95 &4.60 &0.97\\
Ours &$\times$&\textbf{2.24} &\textbf{3.40} &\textbf{0.97} &\textbf{3.50}  &\textbf{4.95}  &\textbf{0.93} &\textbf{2.18} &\textbf{3.20} &\textbf{0.97} &\textbf{2.14} &\textbf{3.37}&\textbf{0.98}\\ 

\bottomrule
\end{tabular}
\end{center}
\end{table*}

\begin{figure}[t]
\begin{center}
\begin{tabular}{cc}
\includegraphics[width=1.6in]{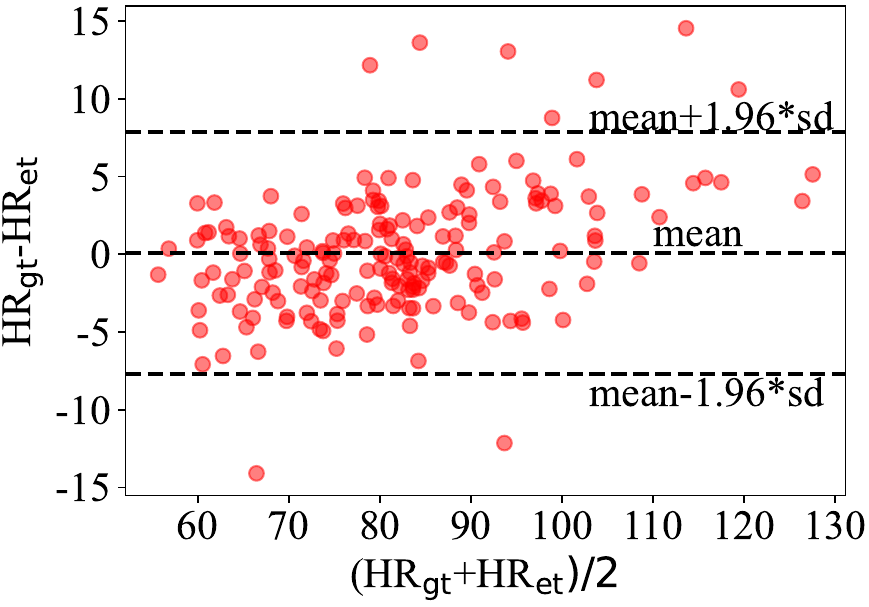} &
\includegraphics[width=1.6in]{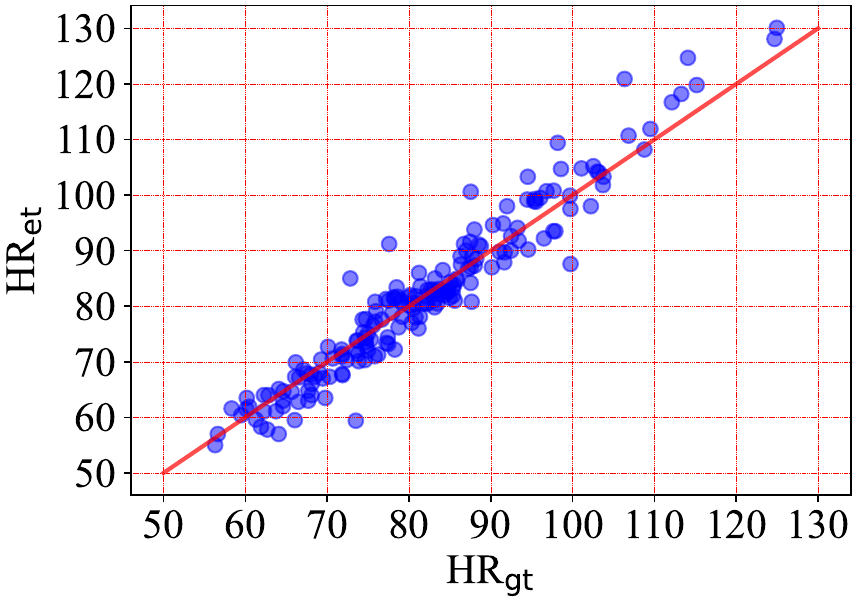} \\
(a) & (b) \\

\end{tabular}
\end{center}
\vspace{-0.1in}
\caption{The Bland-Altman plot (a) and scatter plot (b) show the difference between estimated HR and ground truth HR on the MMVS dataset.}

\label{fig6}
\end{figure}

% \begin{figure}[!t]
%   \centerline{\includegraphics[width=3.5in]{figs/fig7.png}}
%   %\vspace{-0.1in}
%   \caption{One-minute continuous HR estimation results. The blue and red curves
% respectively denotes the ground truth HR the estimated HR.}
%   \label{fig7}
%   \end{figure}
  
Fig.~\ref{fig6} shows the Bland-Altman plot and scatter plot on the MMVS dataset. {HR$_\text{gt}$ and HR$_\text{et}$ represent the HR calculated from the ground truth PPG signal and the estimated rPPG signal, respectively.} Each point indicates an estimation result from one test sample. The x-axis in the Bland-Altman plot denotes the mean value of HR$_\text{gt}$ and HR$_\text{et}$ while the y-axis represents their difference. The top and bottom dashed lines indicate confidence intervals for 95\% limits of agreement. 
% \ZJ{We can observe that the distribution is concentrated at the mean value of 0 bpm and the upper/bottom limit is only 7.9/-7.7 bpm.} 
We can observe that {HR$_\text{gt}$} is well correlated with {HR$_\text{et}$} within a wide range from 50 bpm to 130 bpm. The Pearson's correlation coefficient $r$ between HR$_\text{gt}$ and HR$_\text{et}$ of our approach is 0.94 (Fig.~\ref{fig6}b), which is higher than that of other methods, \eg Meta-rppg \cite{lee_meta-rppg_2020} (0.91), PulseGan \cite{song_pulsegan_2021} (0.93), Physformer \cite{yu_physformer_2022} (0.93), Gideon \etal \cite{gideon_way_2021} (0.93).
%Moreover, we also provide four one-minute continuous HR estimation results in Fig.~\ref{fig7}. The blue and red points respectively represent the ground truth HR and the estimated HR. As shown, the estimated HR are close to the ground truth even facing non-stationary data distributions. Two curves in each figure show similar upward or downward trends, proving the robustness and effectiveness of the proposed approach.the distribution was concentrated at the mean bias of −0.8 bpm and with 95% from −5.1 to 3.2 bpm. There were about 11 points for crossing the (mean ± 1.96 * sd) lines.

\subsubsection{RF and HRV evaluation}

\begin{figure}[t]
\begin{center}
\begin{tabular}{cc}
\includegraphics[width=1.65in]{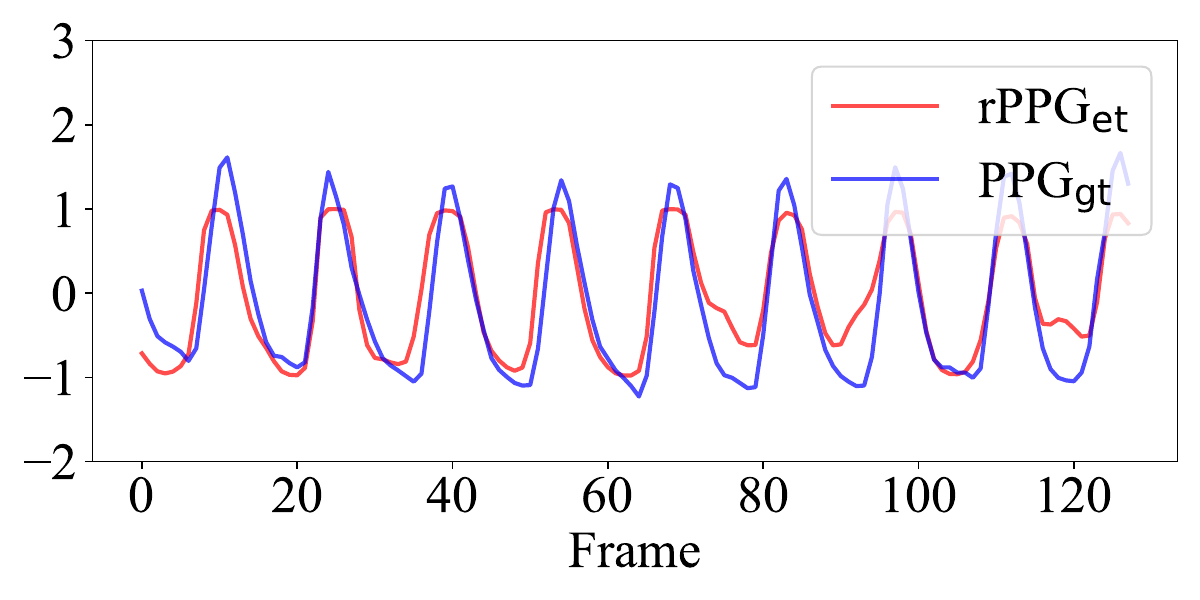} &
\includegraphics[width=1.65in]{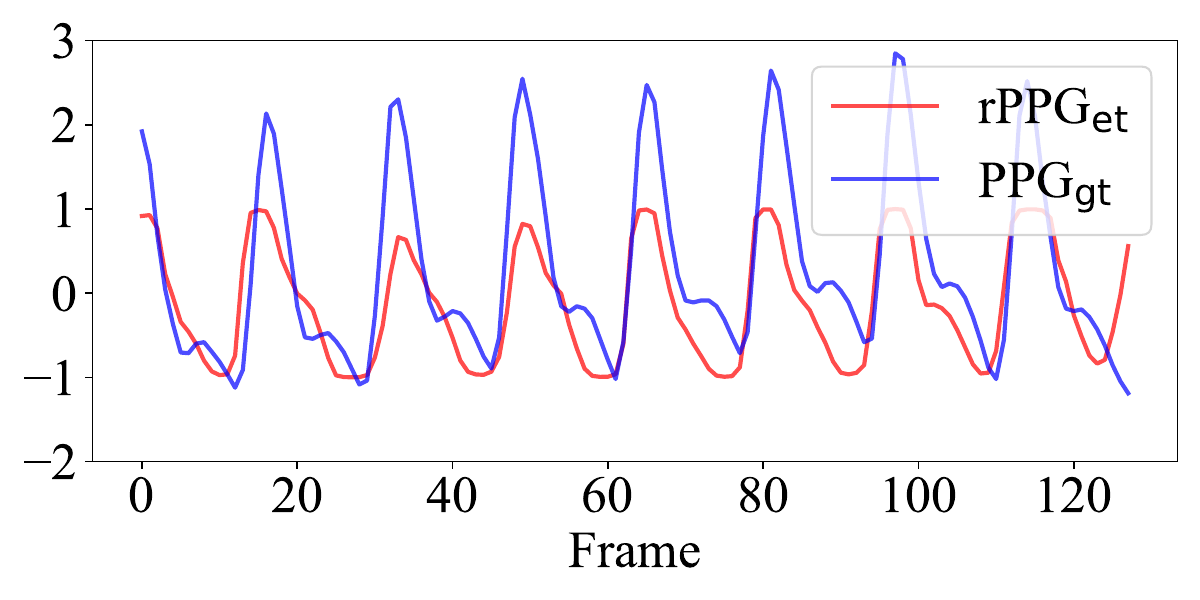} \\

\includegraphics[width=1.65in]{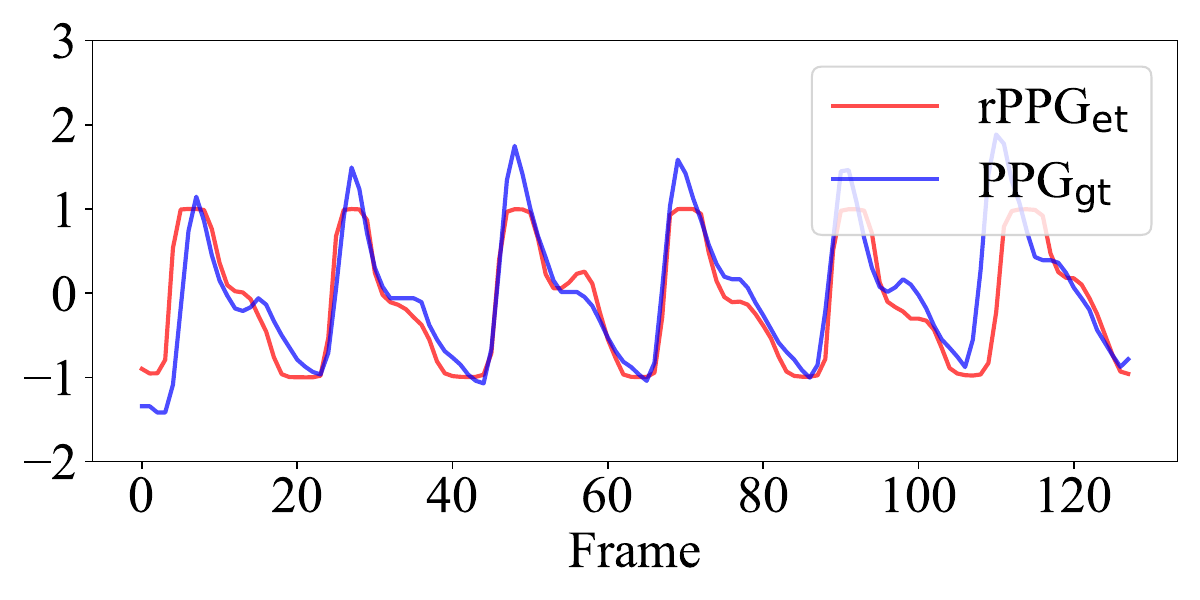} &
\includegraphics[width=1.65in]{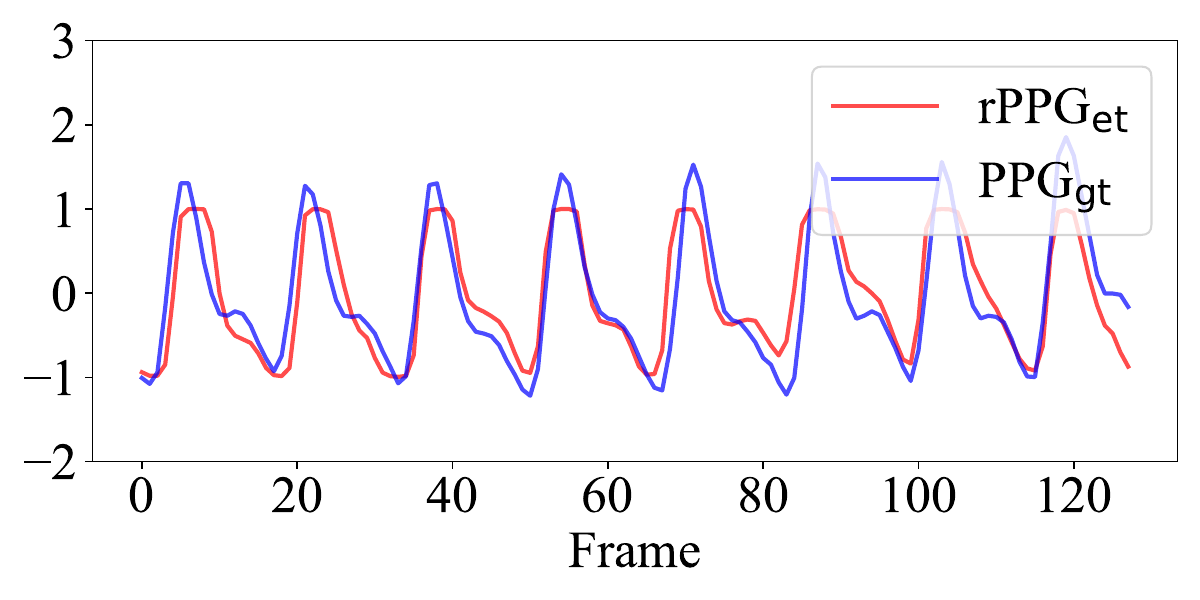} \\
\end{tabular}
\end{center}
\vspace{-0.1in}
\caption{Four examples for the visual comparison between estimated rPPG signals (red curves) and their corresponding ground truth PPG signals (blue curves).}
\label{fig8}
\end{figure}

% \begin{figure}[!t]
%   \centerline{\includegraphics[width=3.5in]{figs/fig7.pdf}}
%   %\vspace{-0.1in}
%   \caption{ } 
%   %Each signal pair has similar interbeat intervals thus we can obtain accurate RF and HRV estimation results.}
%   \label{fig8}
%   \end{figure}

We further conduct experiments for RF and HRV estimation on the UBFC-rPPG dataset. As mentioned in Sec.~\ref{sec4.3}, HRV is represented by its three attributes (LF, HF, LF/HF). Similar to the HR evaluation, we compare our approach with state of the art~\cite{wang_algorithmic_2017,de_haan_robust_2013,verkruysse_remote_2008,niu_video-based_2020,yu_remote_2019,lu_dual-gan_2021,yu_physformer_2022,gideon_way_2021}. The results are shown in Table \ref{table2}. We can see that our approach outperforms all traditional ones and many deep learning ones. For example, as a self-supervised method, we have a comparable performance with the very recent supervised method Physformer \cite{yu_physformer_2022}; moreover, we outperform the self-supervised competitor \cite{gideon_way_2021} by large margins on all metrics.
%All these results indicate that our approach learns more informative features to obtain more promising rPPG signals for RF and HRV estimation.

Fig.~\ref{fig8} shows four estimated rPPG signals and their corresponding ground truth. We can observe that our method predicts rPPG signals with very accurate interbeat intervals compared to ground truth, from which we can obtain robust RF and HRV results.

\subsubsection{Cross-dataset HR evaluation}\label{tab:cross-dataset}

% \begin{figure}[t]
% \begin{center}
% \begin{tabular}{cc}
% \includegraphics[width=1.6in]{figs/fig9a.png} &
% \includegraphics[width=1.6in]{figs/fig9b.png} \\
% (a) & (b) \\
% \end{tabular}
% \end{center}
% \caption{Cross-dataset bland-Altman (a) and scatter (b) plots in terms of HR estimation results on the MMVS dataset.}

% \label{fig9}
% \end{figure}

To evaluate the generalizability of our method, we conduct the cross-dataset experiment among UBFC-rPPG, PURE and MMVS datasets and report the HR estimation results in Table \ref{table4}. Specifically, we train our model, as well as state of the art, \ie \cite{lee_meta-rppg_2020,song_pulsegan_2021,lu_dual-gan_2021,yu_physformer_2022,gideon_way_2021}, on one dataset and test them on another. For example, MMVS$\rightarrow$UBFC-rPPG means training on MMVS while testing on UBFC-rPPG. Ours and \cite{gideon_way_2021} are trained in the self-supervised way while the rests are in the supervised way. The parameter settings for cross-dataset experiments remain the same to original intra-dataset experiments. %\st{UBFC-rPPG$\rightarrow$MMVS vice versa} \ZJ{other columns are defined in the same way.} 
We can see that our method produces very competitive results on par with state of the art supervised methods~\cite{song_pulsegan_2021,lu_dual-gan_2021,yu_physformer_2022}; significantly outperforms the recent self-supervised method~\cite{gideon_way_2021}. This shows the good generalizability of our method 
%to filter out unknown noises and accurately predict rPPG signals from face videos, especially when using it 
in new (unseen) scenarios. 
%Compared to supervised learning, self-supervised learning tends to focus more on generic information than domain-specific information, hence making it more suitable for cross-dataset application. 
%\eg for UBFC-rPPG$\rightarrow$MMVS, it outperforms the second best~\cite{yu_physformer_2022} by decreasing MAE of 0.13 and RMSE of 0.33.

%The Bland-Altman and scatter plots shown in Fig.~\ref{fig9}  demonstrate the good correlation between our estimated HR and ground truth. 

%Our self-supervised learning framework hence shows superior performance on unknown datasets ,  Therefore, we think that self-supervised learning is particularly suitable for our remote physiological measurement task.

\section{Ablation study}\label{sec5}

\begin{table}[!t]
\caption{Ablation study on the frequency modulation block (FMB) and video reconstruction loss in LFA module.}
\vspace{-0.1in}
\label{table5}
\begin{center}
\setlength{\tabcolsep}{1.5mm}
\begin{tabular}{c|ccc|ccc}
\toprule

\multirow{2}*{Method}  &\multicolumn{3}{c|}{UBFC-rPPG}& \multicolumn{3}{c}{MMVS}\\ 

&MAE$\downarrow$  &RMSE$\downarrow$  &$r\uparrow$  & MAE$\downarrow$  &RMSE$\downarrow$  &$r\uparrow$ \\ 
\midrule
Ours  &\textbf{0.58}  &\textbf{0.94} &\textbf{0.99}  &\textbf{2.93} &\textbf{4.16}  &\textbf{0.94}\\
LFA$\rightarrow$VSH &1.27  &1.91  &0.97  &3.21  &4.48  &0.93\\
LFA$\rightarrow$VST \cite{park_bmbc_2020} &0.91  &1.55  &0.97  &3.09  &4.35  &0.93\\
LFA$\rightarrow$VST \cite{park_asymmetric_2021} &0.82  &1.40  &0.98  &3.09  &4.33  &0.93\\
FMB-Linear &1.05  &1.71  &0.97  &3.17 &4.46 &0.93\\
FMB w/o 1D RB &0.80  &1.30  &0.98  &3.03 &4.39 &0.93\\
FMB w/o BiLSTM &0.72  &1.25  &0.99  &3.00 &4.29 &0.93\\
FMB-LSTM &0.66  &1.07  &0.99 &2.98 &4.26 &0.94\\
FMB-GRU &0.68  &1.06  &0.99 &2.98 &4.26 &0.94\\ 
FMB-BiGRU &0.60  &0.97  &0.99 &2.93 &4.19 &0.94\\ 
LFA w/o $\mathcal L_\text{vr}$ &0.88  &1.36  &0.98  &3.04  &4.34  &0.94\\
$\mathcal L_\text{vr} \rightarrow \mathcal L_\text{mr}$ &0.69  &1.11  &0.99  &3.02  &4.30  &0.93\\
% FRCO w/o $l_{frl}$ &2.86  &4.03  &0.96  &3.06  &4.24  &0.94\\
% FRCO w/o $l_{pro}$ &2.64  &3.82  &0.97  &3.03  &4.19  &0.94\\

\bottomrule
\end{tabular}
\end{center}
\end{table}

We conduct ablation study from five aspects: LFA module, REA module, frequency-inspired losses, temporally neighboring videos and spatial augmentation. The results for HR estimation are reported on UBFC-rPPG and MMVS datasets.

\subsection{Learnable frequency augmentation (LFA) module}\label{sec5.1}
%\miaojing{use subsubsections for the below titles.}
\subsubsection{LFA \vs video shrinking/stretching}\label{sec5.1.1}
Table \ref{table5} presents the ablation study on the proposed LFA module.
% FAM, frequency ratio loss $l_{frl}$ and projection loss $l_{pro}$. 
% \st{First,} \ZJ{LFA module aims to generate multiple negative samples with various rPPG signal frequencies, and this goal can also be theoretically achieved by using video shrinking and stretching operations. Therefore we first verify the advantages of the LFA module over these operations. }
Instead of learning to generate negative samples in the network, we can also leverage video shrinking (VSH) and video stretching (VST) to achieve this. We use LFA$\rightarrow$VSH and LFA$\rightarrow$VST to indicate the two variants in Table~\ref{table5} and compare them to our LFA.

%instead, we \ZJ{leverage video shrinking and video stretching operation for negative sample generation, respectively. 
Specifically, for video shrinking, we follow \cite{gideon_way_2021} to reduce the frame number of the input video and generate one negative sample; for video stretching, the most common way is through video frame interpolation; we select two state of the art video frame interpolation methods, \ie Bilateral Motion Estimation with Bilateral Cost Volume (BMBC) \cite{park_bmbc_2020} and Asymmetric Bilateral Motion Estimation (ABME) \cite{park_asymmetric_2021}, to respectively stretch the video and generate negative samples.

%\cite{park_bmbc_2020} and LFA$\rightarrow$VSH \cite{park_asymmetric_2021}}. 
In Table \ref{table5}, we can observe severe performance degradation by replacing LFA with VSH or VST.
%\st{, which justifies the advantage of LFA module for obtaining negative samples with more diverse rPPG signal frequencies.} 
The reason is two-fold: 1) in LFA$\rightarrow$VSH, the frame number of the shrunk video will be less than that of the original video, which means the number of sampling points of the extracted rPPG signal will also be reduced. Insufficient number of sampling points in the signal can bring in spurious components on the power spectral PSD analysis \cite{mendez-diaz_power_2013}, which is not desired as we need precise PSD in our frequency-inspired losses; 2) in LFA$\rightarrow$VST, the rPPG signal frequency of the stretched video can only be lower than that of the original video (the number of signal waves per unit of time is decreased). This would lead to poor signal frequency diversity, hence impeding the model training. Besides, the majority of video interpolation methods pay more attention to recovering smooth motions than restoring skin color changes. The original periodicity or quasi-periodicity of rPPG waveforms is likely to be destroyed after adopting the video interpolation methods. Overall, video shrinking/stretching is not a good option for generating negative samples.

% The above observations}
% \st{which justifies} \ZJ{justify} the advantage of LFA module for obtaining negative samples with more diverse rPPG signal frequencies. \ZJ{Moreover, LFA module can overcome the weakness of video shrinking to obtain sufficient number of frames for model optimization, and avoid performing video stretching to destroy the original periodicity or quasi-periodicity of rPPG waveforms. Thus it is more effective on frequency augmentation.}

%\st{our contribution of contrasting each positive sample with multiple frequency-modulated negative samples to improve the representation learning ability of MORE.} 

% Compared with \cite{gideon_way_2021}, which can only manually enlarge the frequency of rPPG signal contained in anchor sample, our FAM can learn to automatically transform rPPG signals to the target higher/lower frequency versions. The increased frequency diversity can help MORE to capture more informative and useful information. 

% Moreover, after we remove $l_{frl}$ and $l_{pro}$, the performance also clearly declines on these two datasets. These results indicate that $l_{frl}$ is effective to enhance the frequency augmentation effects of FAM, and $l_{pro}$ can help the network to maintain signal periodicity over a long period of time (Fig.~\ref{fig8} demonstrates the satisfactory periodicity of the estimated signals).

\begin{table}[!t]
\caption{Ablation study on the multi-scale architecture in LFA module.}
\vspace{-0.1in}
\label{table6}
\begin{center}
\setlength{\tabcolsep}{1.5mm}
\begin{tabular}{c|ccc|ccc}
\toprule
\multirow{2}*{Method}  &\multicolumn{3}{c|}{UBFC-rPPG}& \multicolumn{3}{c}{MMVS}\\ 

&MAE$\downarrow$  &RMSE$\downarrow$  &$r\uparrow$  & MAE$\downarrow$  &RMSE$\downarrow$  &$r\uparrow$ \\ 
\midrule
LFA-S1  &0.83  &1.32  &0.98  &3.04  &4.33  &0.94\\
LFA-S2 &0.66  &1.06  &0.99  &2.99  &4.24  &0.94\\
LFA-S3 &\textbf{0.58}  &\textbf{0.94} &\textbf{0.99}  &\textbf{2.93} &4.16  &\textbf{0.94}\\
LFA-S4 &0.58  &0.95  &0.99  &2.96  &\textbf{4.15}  &0.94\\

\bottomrule
\end{tabular}
\end{center}
\end{table}

\subsubsection{Multi-scale in LFA}\label{sec5.1.2}
The LFA module performs frequency transformation on three scales (S3). We also experiment with one scale (S1), two scales (S2) and four scales (S4) in Table \ref{table6}. They are denoted as LFA-S1, LFA-S2 and LFA-S4, respectively. We can observe that LFA-S3 performs in general the best. LFA-S4 performs very close to LFA-S3 yet with additional computation. LFA-S1/S2 performs clearly inferior to LFA-S3. Overall, the three-scale version appears to be sufficient to describe different levels of details for the input. This is our default setting. 
% \ZJ{since three-scale pyramid structure has been demonstrated to be sufficient to extract different levels of details of the input in }

%feature scales have better performance and peaking at LFA-S3. 

%The frequency transformation on three scales are all beneficial to enhance the overall performance. 

% \begin{figure}[t]
% \begin{center}
% \begin{tabular}{cc}
% \includegraphics[width=1.65in]{figs/fig10a.png} &
% \includegraphics[width=1.65in]{figs/fig10b.png} \\
% \scriptsize (a) MAE on UBFC-rPPG & \scriptsize (b) RMSE on UBFC-rPPG\\
% \includegraphics[width=1.65in]{figs/fig10c.png} &
% \includegraphics[width=1.65in]{figs/fig10d.png} \\
% \scriptsize (c) MAE on MMVS & \scriptsize (d) RMSE on MMVS \\
% \end{tabular}
% \end{center}
% \caption{The effect of different number of negative samples in LFA. The results are reported on the UBFC-rPPG and MMVS datasets.}
% \label{fig10}
% \end{figure}

\subsubsection{Feature modulation block in LFA} \label{sec5.1.3}
We further verify the design of the feature modulation block (FMB) in the LFA module. As mentioned in Sec.~\ref{sec3.2.1}, the key part of FMB is the 1D RB and BiLSTM for non-linear signal frequency transformation. A linear transformation is not adequate. For instance, we can replace 1D RB + BiLSTM with three 1D Convs, which we call it FMB-Linear in Table \ref{table5}. We see that the performance {drops} substantially by FMB-Linear, attesting that non-linear functions are important for frequency transformation. 

Furthermore, we use FMB w/o 1D RB and FMB w/o BiLSTM to denote the FMB without 1D RB or BiLSTM. We can observe that both 1D RB and BiLSTM contribute to the final result. 1D RB is responsible for the signal's local transformation followed by a non-linear activation function (ReLu). BiLSTM dedicates to the signal's global transformation along the temporal dimension. Alternatively, we can replace BiLSTM with other RNN units, such as LSTM, GRU and BiGRU. We name these variants as FMB-LSTM, FMB-GRU and FMB-BiGRU in Table \ref{table5}: they show competitive performance but our FMB with BiLSTM is the best.

\subsubsection{Video reconstruction loss in LFA} \label{sec5.1.4}
We utilize video reconstruction loss $\mathcal L_\text{vr}$ to reduce color differences between negative samples and the original sample. If we remove $\mathcal L_\text{vr}$, the result (LFA w/o $\mathcal L_\text{vr}$ in Table \ref{table5}) will be much inferior. %$\mathcal L_\text{vr}$ helps LFA \st{retain perceptual details in augmented samples while only modulating their underlying signal frequencies} \ZJ{modulate underlying signal frequencies of original samples without changing their colors 
%excessively. 
As mentioned in Sec.~\ref{sec3.2.2}, an alternative to $\mathcal L_\text{vr}$ is to regulate the modulation vector, \ie $\mathcal L_\text{mr} =  {\textstyle \frac{1}{k} \sum_{i=1}^{k}}\|\mathbf m_{i} -1\|_2$. We report the result for this variant in Table \ref{table5}: $\mathcal L_\text{vr} \rightarrow \mathcal L_\text{mr}$, it works by improving upon LFA w/o $\mathcal L_\text{vr}$ yet it is no better than Ours with $L_\text{vr}$.  
%we can also achieve similar effect by directly constraining the modulation vector. 
%We term this variant as $\mathcal L_\text{vr} \rightarrow \mathcal L_\text{m}$, and our original model performs better than it.
%Without $\mathcal L_\text{vr}$, the LFA module can easily change spatial features of anchor sample, leading to poor model performance.}

\subsubsection{Number of negative samples in LFA} \label{sec5.1.5}
%The training batch size of conventional self-supervised learning approaches is an important factor that affects the encoder performance. This is because a larger batch size can bring in more negative samples to learn their representation dissimilarities.
% , thus the encoder can be trained to have a stronger feature representation ability. 
In the training phase, we found that the number of negative samples affects the model performance. We vary this number $k$ from 1 to 6 and report the result in Table \ref{table7}. Our default setting ($k = 4$) appears to be the best.
% \ZJ{We can observe that the model performance is similar when $k$ is bigger than 4 (our default setting).}
% \ZJ{following \cite{kim_self-supervised_2021}} 
% \ZJ{Note that in the variant Ours (k=1), besides generating only one negative sample, we also follow  \cite{gideon_way_2021} to degrade our method to generate only one positive sample for self-supervised training, that means the anchor sample is also regarded as another positive sample so as to apply losses between their signals.}
%A bigger $k$ shows better performance but also brings in additional cost. Considering the trade-off between performance and computation, 

\begin{table}[!t]
\caption{Ablation study on different number of negative samples in LFA module.}
\vspace{-0.1in}
\label{table7}
\begin{center}
\begin{tabular}{c|ccc|ccc}
\toprule
\multirow{2}*{Method}  &\multicolumn{3}{c|}{UBFC-rPPG}& \multicolumn{3}{c}{MMVS}\\ 

&MAE$\downarrow$  &RMSE$\downarrow$  &$r\uparrow$  & MAE$\downarrow$  &RMSE$\downarrow$  &$r\uparrow$ \\ 
\midrule
Ours ($k = 1$)  &1.10  &1.87  &0.97  &3.10  &4.28  &0.93\\
Ours ($k = 2$)  &1.03  &1.66  &0.97  &3.07  &4.35  &0.94\\
Ours ($k = 3$) &0.75  &1.27 &0.98  &3.00  &4.23  &0.94\\
Ours ($k = 4$) &\textbf{0.58}  &0.94 &\textbf{0.99}  &\textbf{2.93}  &4.16  &\textbf{0.94}\\
Ours ($k = 5$) &0.58  &0.96  &0.99  &2.93  &\textbf{4.15}  &0.94\\
Ours ($k = 6$) &0.59  &\textbf{0.92}  &0.99  &2.96  &4.15  &0.94\\

\bottomrule
\end{tabular}
\end{center}
\end{table}

\begin{table}[!t]
\caption{Ablation study on the proposed REA module.}
\vspace{-0.1in}
\label{table8}
\begin{center}
\setlength{\tabcolsep}{1.3mm}
\begin{tabular}{c|ccc|ccc}
\toprule
\multirow{2}*{Method}  &\multicolumn{3}{c|}{UBFC-rPPG}& \multicolumn{3}{c}{MMVS}\\ 

&MAE$\downarrow$  &RMSE$\downarrow$  &$r\uparrow$  & MAE$\downarrow$  &RMSE$\downarrow$  &$r\uparrow$ \\ 
\midrule
Ours &\textbf{0.58}  &\textbf{0.94} &\textbf{0.99}  &\textbf{2.93} &\textbf{4.16}  &\textbf{0.94}\\
REA$\rightarrow$GRA &1.16  &1.85  &0.97 &3.29  &4.61  &0.93\\ 
REA w/o RA &1.02  &1.69  &0.97  &3.13  &4.40  &0.93\\ 
RA$\rightarrow$SA &0.62  &0.95  &0.99  &2.96  &4.22  &0.94\\ 
$\mathcal{G}\rightarrow \text{avg}$ &1.12 &1.89 &0.97 &3.23 &4.56 &0.93\\ 
$\mathcal{G}\rightarrow \mathcal{G}_{spat}$ &0.76 &1.26 &0.98 &3.07 &4.41 &0.94\\ 
\bottomrule
\end{tabular}
\end{center}
\end{table}

\begin{table}[!t]
\caption{Ablation study on the number of experts in REA module.}

\label{table11}
\begin{center}
\setlength{\tabcolsep}{1.3mm}
\begin{tabular}{c|ccc|ccc}
\toprule
\multirow{2}*{Method}  &\multicolumn{3}{c|}{UBFC-rPPG}& \multicolumn{3}{c}{MMVS}\\ 

&MAE$\downarrow$  &RMSE$\downarrow$  &$r\uparrow$  & MAE$\downarrow$  &RMSE$\downarrow$  &$r\uparrow$ \\ 
\midrule
Ours ($L = 2$)  &0.75  &1.15  &0.98  &3.17  &4.48  &0.93\\
Ours ($L = 4$) &0.61  &0.98  &0.99  &2.97 &4.19 &0.94\\
Ours ($L = 9$) &0.58  &\textbf{0.94} &\textbf{0.99}  &\textbf{2.93} &\textbf{4.16}  &\textbf{0.94}\\
Ours ($L = 16$) &\textbf{0.57}  &0.94 &0.99  &2.95  &4.16  &0.94\\
Ours ($L = 25$) &0.58  &0.95 &0.99  &2.94  &4.18  &0.94\\
\bottomrule
\end{tabular}
\end{center}
\end{table}

\subsection{Local rPPG expert aggregation (REA) module}\label{sec5.2}
We investigate the effectiveness of the REA module in Table \ref{table8}. The first variant is to not use the local rPPG experts but simply extract the rPPG signal from the entire face. The REA module degrades to one 3D ResNet-10 block at the beginning followed by one 3D GAP and one 1D Conv for rPPG estimation. We denote this variant as a global rPPG estimation (GRA) module, \ie REA$\rightarrow$GRA, in Table~\ref{table8}. % denotes the  variant in which we only keep the beginning 3D ResNet in the the current REA module, and the output of 3D ResNet-10 is directly fed into a . 
We can observe that the MAE increases by 0.58 and the RMSE increases by 0.91 on UBFC-rPPG. This shows that encoding complementary pulsation information from different face regions helps improve the rPPG estimation. Below we study the importance of the region-attention (RA) block and spatio-temporal gating net in the REA module.

\subsubsection{Region-attention block in REA} \label{sec5.2.1}
Referring to Fig.~\ref{fig4}b, we can replace the RA block with a 3D RB in the REA module. We denote this by REA w/o RA in Table~\ref{table8}: the MAE increases to 1.02, RMSE increases to 1.69, and $r$ decreases to 0.97 on the UBFC-rPPG dataset. The proposed RA block is more effective than a simple 3D RB in finding the pulsation-sensitive area in a local region. Alternatively, we follow \cite{yu_remote_2019} to use an additional skin segmentation branch after the 3D ResNet-10 block. This branch produces a seg-attention (SA) map, which we use to replace our region-attention map and is multiplied back to the encoded feature. 
%to assign high/low weights on skin/non-skin regions, next we remove the region-attention blocks from REA module. 
We denote this by RA$\rightarrow$ SA. It performs better than REA w/o RA but is still inferior to RA. SA map assigns weights to pixels according to their likelihood of being face skins; our RA map instead assigns weights to pixels according to their sensitivities to pulsation changes. RA can capture more discriminative physiological clues from face skins for rPPG estimation.
% we both assign weights on the features maps in spatial domain to enhance information from skins and suppress information from irrelevant backgrounds.

% \begin{figure}[t]
% \begin{center}
% \begin{tabular}{c}
% \includegraphics[width=2.7in]{figs/fig_major5-1.pdf}\\
% \includegraphics[width=2.7in]{figs/fig_major5-2.pdf} \\
% \end{tabular}
% \end{center}
% \vspace{-0.1in}
% \caption{\ZJ{Visualization of the weights assigned to different experts at different moments. The brighter rectangular indicates the larger weight.}}
% \label{fig16}
% \end{figure}

\subsubsection{Spatio-temporal gating net in REA} \label{sec5.2.2}
The spatio-temporal gating net $\mathcal{G}$ assigns weights to local rPPG experts for their aggregation. If we remove it from the REA module but take the average of multiple experts, the MAE and RMSE will be significantly increased for HR estimation (see Table~\ref{table8}: $\mathcal{G}\rightarrow avg$). 
%that it only has XXX MAE and XXX RMSE on MMVS. 
This supports our claim that the distribution of blood vessels varies over face regions. Our gating net treats these regions differently in rPPG estimation. Next, we adjust $\mathcal{G}$ to assign one scalar weight to one expert (corresponding to a spatial region) while ignoring the temporal change in the signal. We denote this by {$\mathcal{G}\rightarrow \mathcal{G}_{spat}$} in Table~\ref{table8}: one can clearly see the performance drop by this variant. 
%($L$ scalars rather than $L$ vectors with dimension {$1 \times T$}) to $L$ experts. We call it $\mathcal{G}\rightarrow \mathcal{G}_{hard}$ and it is inferior to $\mathcal{G}$. 
This validates our motivation that each expert signal should be assigned different weights at different moments (see Sec.~\ref{sec:gatingnet}). 

% \ZJ{We also visualize the weights assigned to different experts at different moments by the gating net in Fig.~\ref{fig16}. It is interesting to find that the experts of forehead, nose, left cheek and right cheek regions are often assigned with large weights. Coincidentally, these regions are reported to have abundant capillaries beneath the skins to reflect the periodic change of blood volume \cite{yu_physformer_2022}. The results show that our gating net can distinguish them with other background or pulsation-insensitive skin regions, and assign their corresponding experts with larger weights to obtain accurate rPPG signal estimation results.}

%Some regions may reflect rPPG signal better during the systole while some are more sensitive during the diastole. \miaojing{discuss this earlier} This is indeed the designing spirit of our spatio-temporal gating net $\mathcal{G}$. 
%\miaojing{hard weights can also be on pixel-level}
% To test the effectiveness of these gating nets, we remove them from LTMoE and obtain interpolated features by 1) averaging the experts directly, 2) concatenating the experts and then leverage a 1\times1\times1 3D convolutional layer for dimensionality reduction,

\subsubsection{Number of experts in REA} \label{sec5.2.3}
%The number of  also affects the model performance. 
We vary the number of local rPPG experts $L$ from 2, 4 ($2 \times 2$), 9 ($3 \times 3$), 16 ($4 \times 4$), and 25 ($5 \times 5$) in the REA module and show the performance in Table \ref{table11}. Our default setting ($L=9$) appears to be the best. Yet, the performance difference for $L$ from 4 to 25 is not big.
% \ZJ{inspired by \cite{yu_physformer_2022}}

%denotes we divide the entire face into $9$ partitioned regions and generate $9$ local rPPG experts accordingly. 
%Then the spatio-temporal gating net outputs $9$ vectors to aggregate experts for rPPG estimation. Other variants in Table \ref{table11} are defined in the same way. 
%We can observe that Ours ($L=9$), Ours ($L=16$) and Ours ($L=25$) have comparable performance. In order to reduce the computation cost, we select Ours ($L=9$) as default.

\begin{table}[!t]
\caption{Ablation study on the frequency-inspired losses.}
\label{table9}
\begin{center}
\setlength{\tabcolsep}{1.5mm}
\begin{tabular}{c|ccc|ccc}
\toprule
\multirow{2}*{Method}  &\multicolumn{3}{c|}{UBFC-rPPG}& \multicolumn{3}{c}{MMVS}\\ 

&MAE$\downarrow$  &RMSE$\downarrow$  &$r\uparrow$  & MAE$\downarrow$  &RMSE$\downarrow$  &$r\uparrow$ \\ 
\midrule
Ours  &0.58  &\textbf{0.94} &\textbf{0.99}  &\textbf{2.93} &4.16  &\textbf{0.94}\\
Ours w/o $\mathcal L_\text{fc}$ &2.13  &3.09  &0.83  &5.00  &6.74  &0.85\\
Ours w/o $\mathcal L_\text{fr}$ &0.89  &1.39  &0.98  &3.07  &4.35  &0.94\\
Ours w/o $\mathcal L_\text{fa}$ &0.67  &1.08  &0.99  &3.04  &4.30  &0.94\\
$\mathcal L_\text{fr} \rightarrow \mathcal L_\text{fr-single}$ &0.66  &1.08  &0.99  &3.00  &4.24  &0.94\\
$\mathcal L_\text{fa} \rightarrow \mathcal L_\text{fa-single}$ &0.60  &0.98  &0.99  &3.00  &4.26  &0.94\\
Ours-expand &\textbf{0.55}  &0.96  &0.99  &2.94  &\textbf{4.14}  &0.94\\

\bottomrule
\end{tabular}
\end{center}
\end{table}

\subsection{Frequency-inspired losses}\label{sec5.3}
To validate the effectiveness of our proposed frequency contrastive loss $\mathcal L_\text{fc}$, frequency ratio consistency loss $\mathcal L_\text{fr}$, cross-video frequency agreement loss $\mathcal L_\text{fa}$, we ablate them {one by one} in our framework and show the results in Table~\ref{table9}. We observe a clear performance drop by removing any of them. 
%ratio consistency loss $\mathcal L_\text{fr}$ and cross-video frequency agreement loss $\mathcal L_\text{fa}$, we remove either one and re-train the model. Table \ref{table7} 
 
%for FISL w/o $\mathcal L_\text{fr}$ and FISL w/o $\mathcal L_\text{fa}$.
\subsubsection{Frequency contrastive loss $\mathcal L_\text{fc}$} \label{sec5.3.1}
Without using $\mathcal L_\text{fc}$, the MAE and RMSE for HR evaluation are substantially increased while $r$ is substantially decreased, \eg 5.00, 6.74 and 0.85 on MMVS. $\mathcal L_\text{fc}$ enforces the signal frequency similarities among positive samples and dissimilarities {between positive and} negative samples. It optimizes the model embedding space to be discriminative to skin color changes. 

\begin{figure}[t]
\begin{center}
\begin{tabular}{cc}
\includegraphics[width=1.65in]{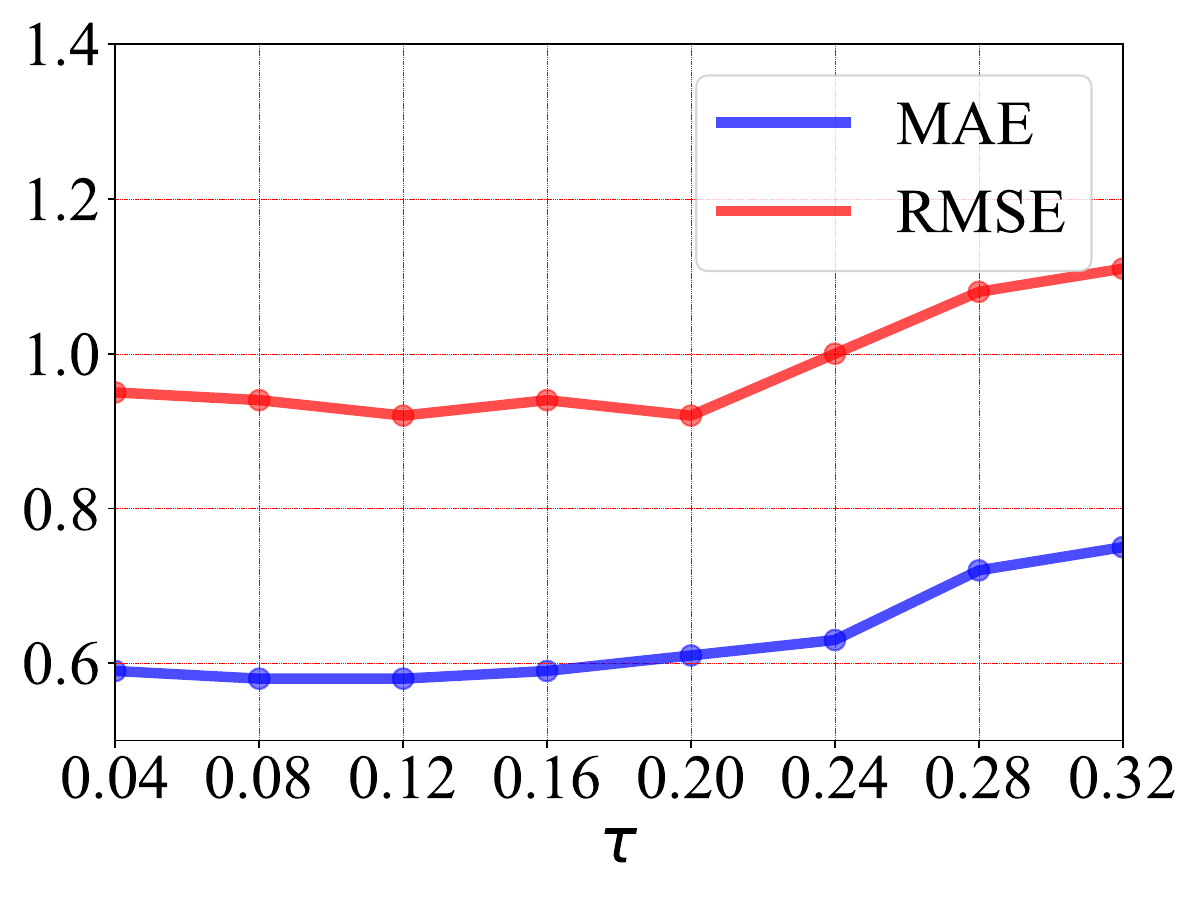} &
\includegraphics[width=1.65in]{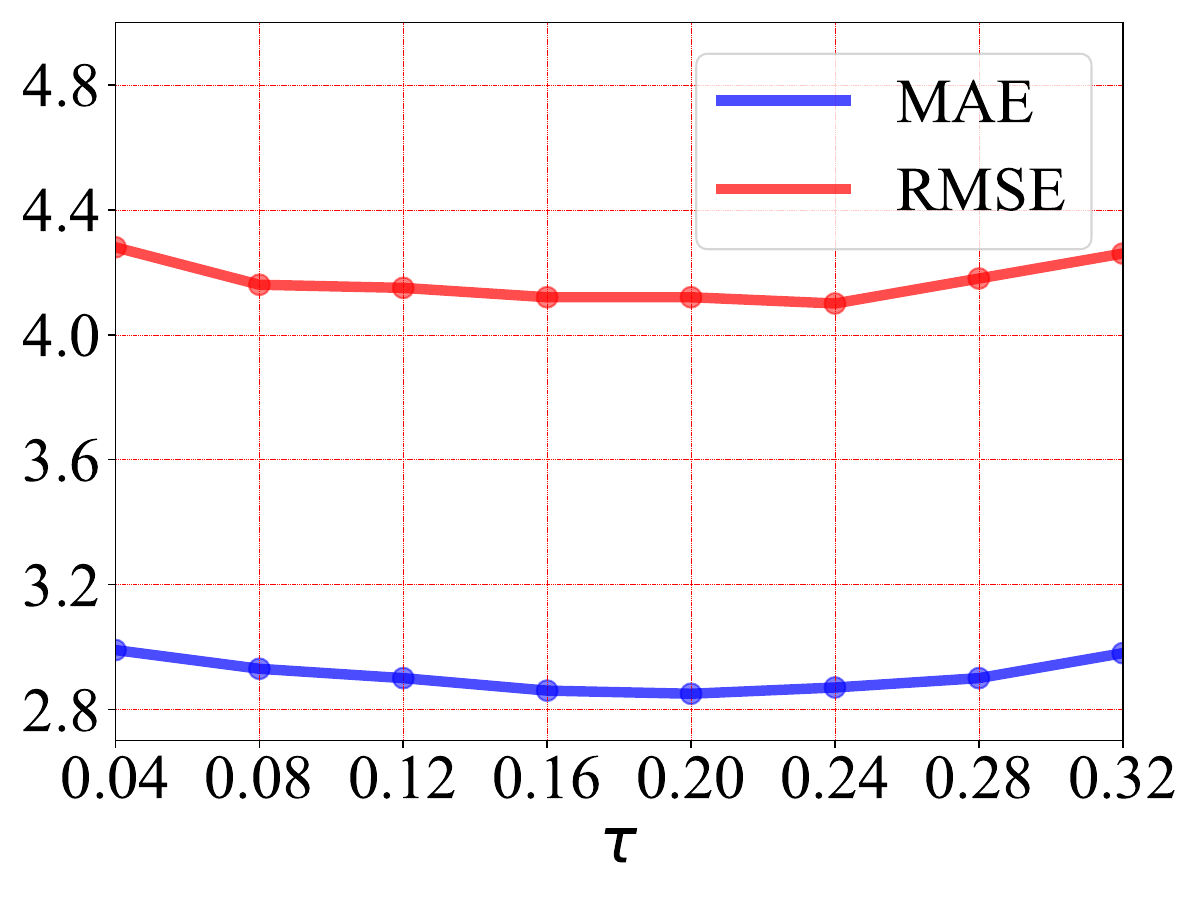} \\
\scriptsize (a) UBFC-rPPG & \scriptsize (b) MMVS\\
\end{tabular}
\end{center}
\vspace{-0.1in}
\caption{The effect of temperature parameter ($\tau$) in our frequency contrastive loss.}
\label{fig17}
\end{figure}

We further vary the temperature parameter $\tau$ in $\mathcal L_\text{fc}$ from 0.04 to 0.32 to explore its influence on the model performance. The MAE and RMSE on HR estimation are shown in Fig.~\ref{fig17}. We can observe that, for both datasets, the performance of our default setting of $\tau=0.08$ is close to the best, and also does not fluctuate much in the neighborhood. Hence, we recommend sticking with the default setting for the sake of model's generalizability. The temperature plays a role in controlling the strength of penalties on hard negative samples~\cite{wang_understanding_2021}. A small temperature tends to penalize more on hard samples. In Fig.~\ref{fig17} we can see that the best $\tau$ on the MMVS dataset is bigger than that of UBFC-rPPG. This is consistent with our experimental finding in Table \ref{table1}: results on UBFC-rPPG are generally better than on MMVS. This indicates UBFC-rPPG is an easier dataset compared to MMVS. A small temperature on UBFC-rPPG will focus the model on distinguishing hard negative samples. While for MMVS, a large temperature is more suitable to let the model distinguish the majority of negative samples as they are already quite difficult.

%Second, we can see that the best $\tau$ on the MMVS dataset is bigger than that of UBFC-rPPG. This indicates UBFC-rPPG is an easier dataset compared to MMVS because the temperature plays a role in controlling the strength of penalties on hard negative samples\cite{wang_understanding_2021}. A smaller temperature on UBFC-rPPG will focus the model on distinguishing harder negative samples. While for MMVS, a large temperature is more suitable to let the model distinguish the majority of negative samples as they are already quite difficult. This is consistent with our experimental finding in Table \ref{table1}: results on UBFC-rPPG are generally better than on MMVS.}

\begin{figure}[t]
\begin{center}
\begin{tabular}{cc}
\includegraphics[width=1.65in]{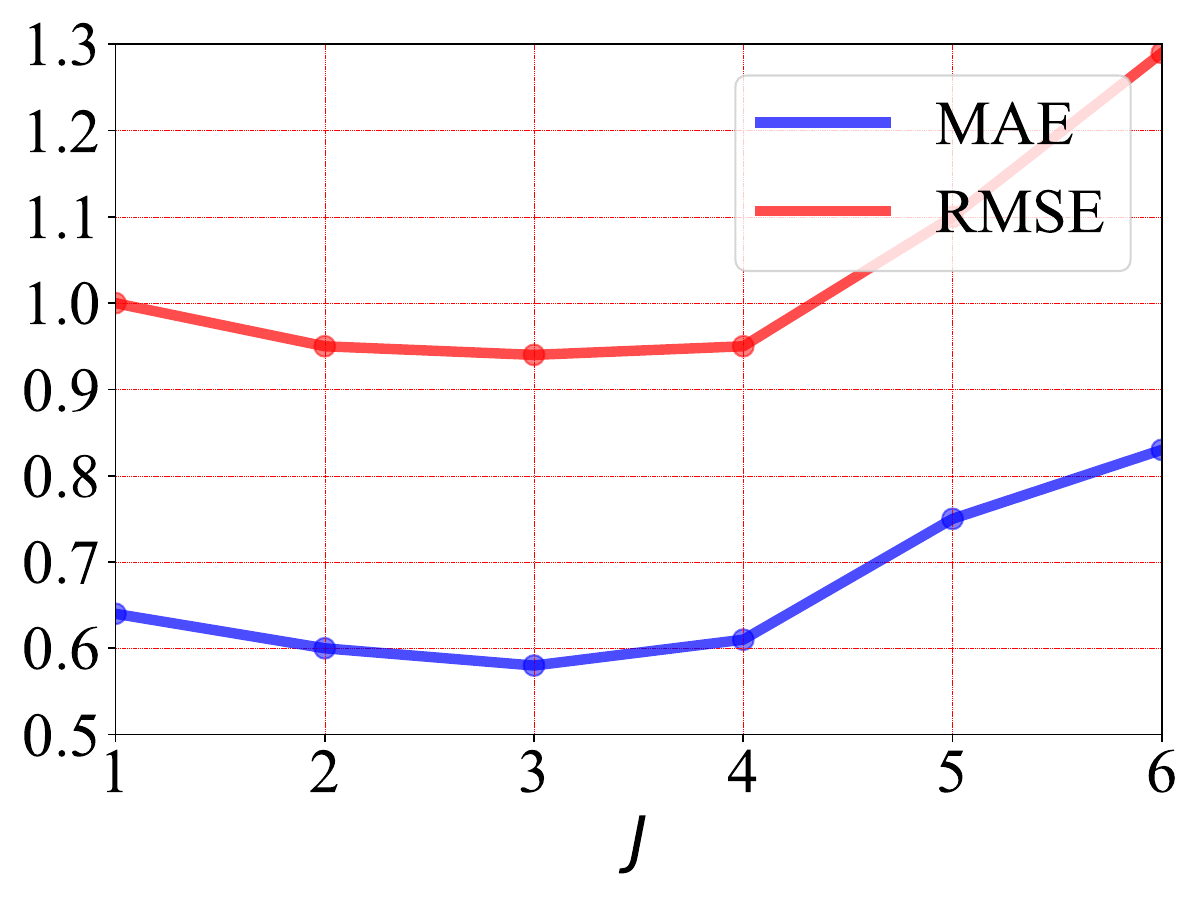} &
\includegraphics[width=1.65in]{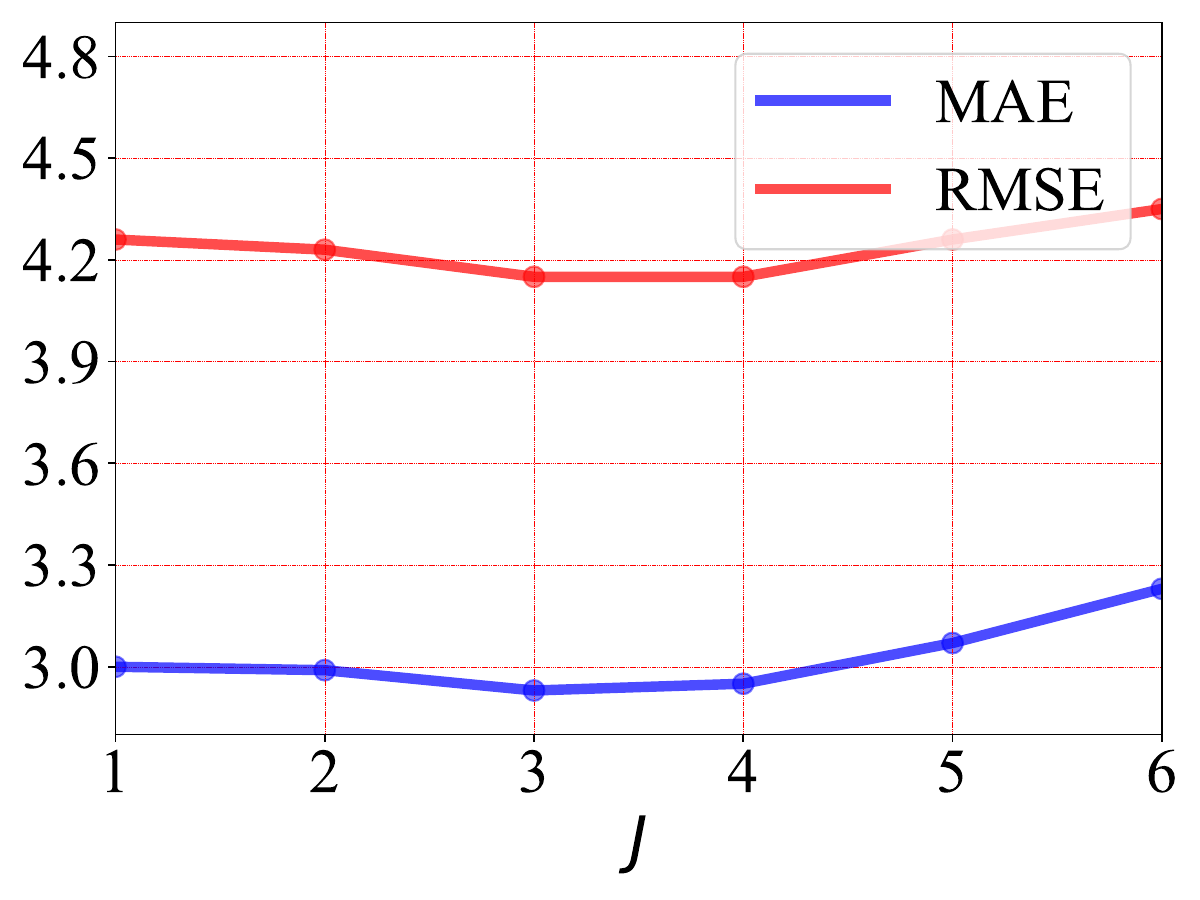} \\
\scriptsize (a) UBFC-rPPG & \scriptsize (b) MMVS\\
\end{tabular}
\end{center}
\vspace{-0.1in}
\caption{Parameter variation on the number of video clips.}
\label{fig12}
\end{figure}

\begin{figure}[t]
\begin{center}
\begin{tabular}{cc}
\includegraphics[width=1.65in]{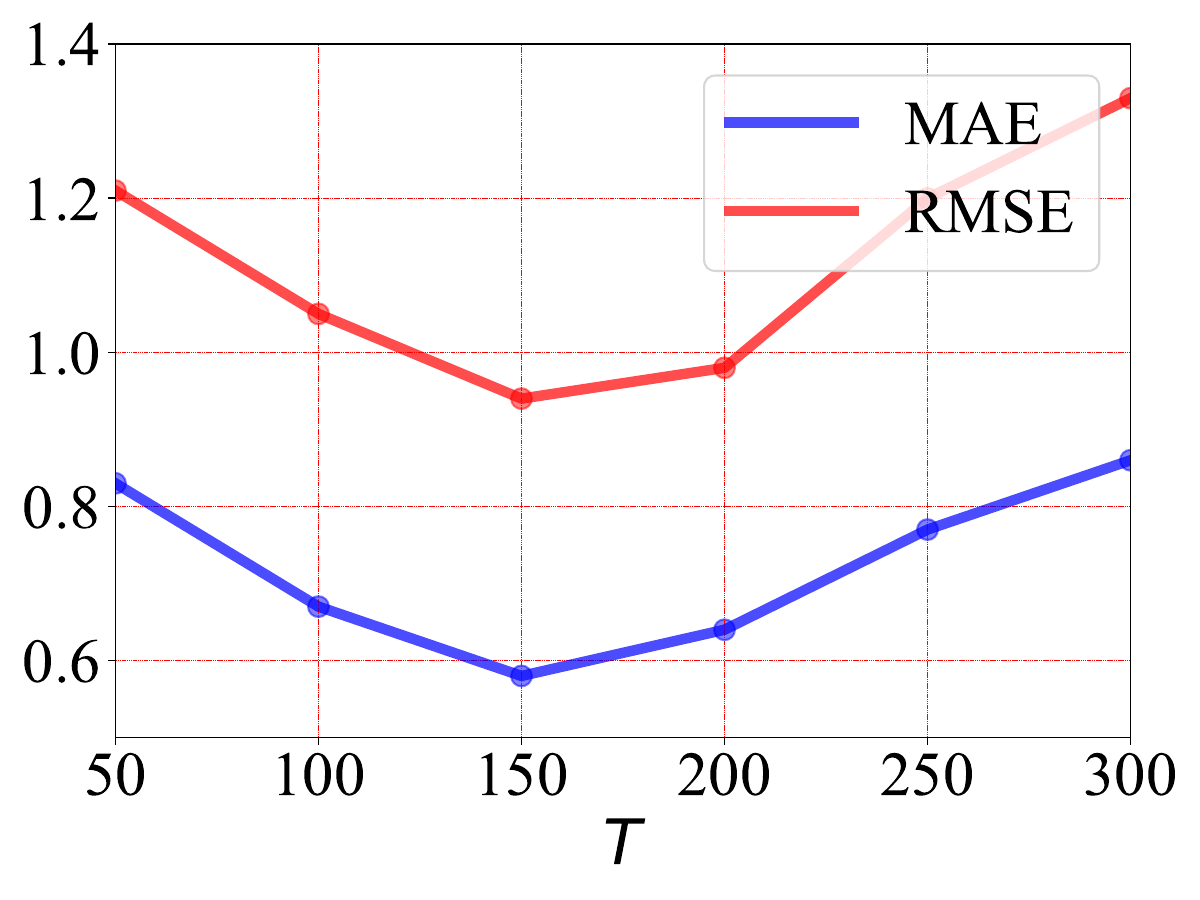} &
\includegraphics[width=1.65in]{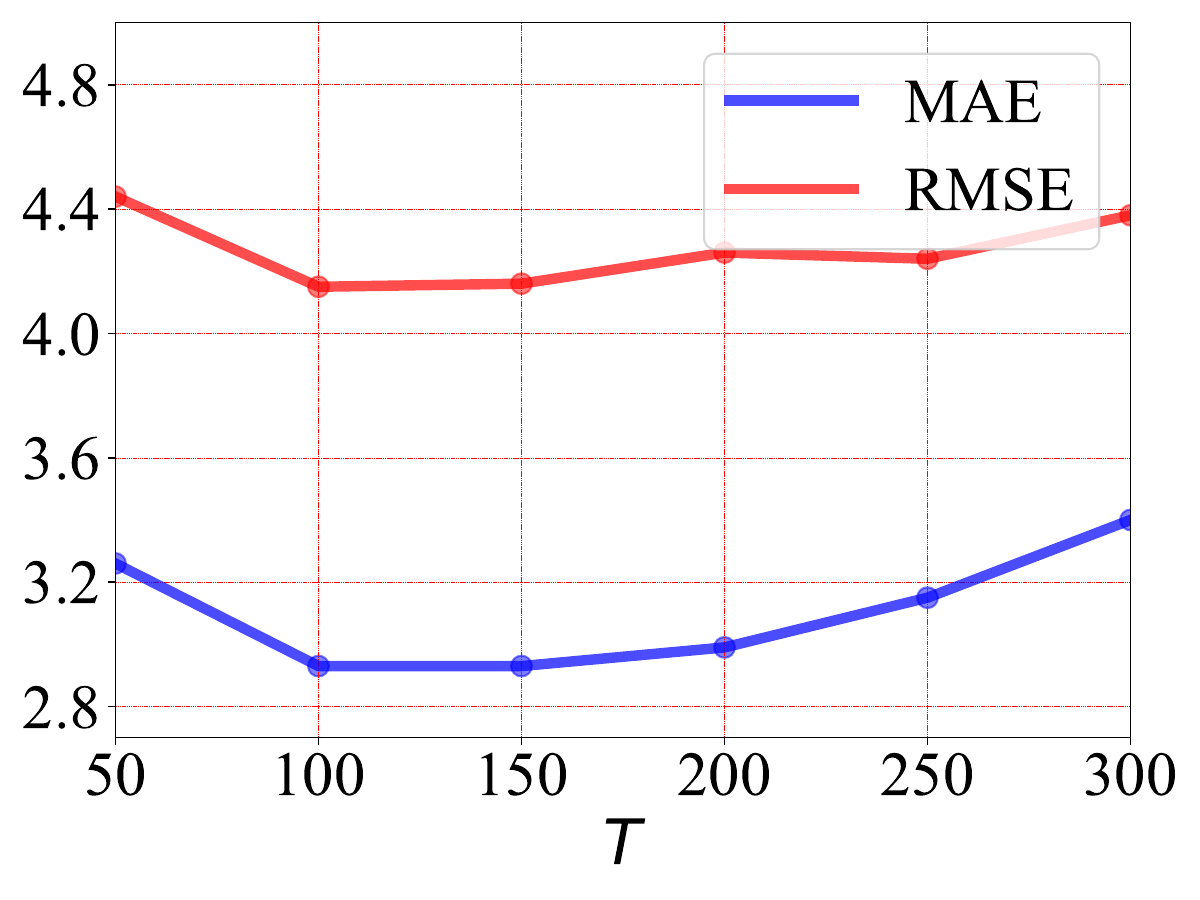} \\
\scriptsize (a) UBFC-rPPG & \scriptsize (b) MMVS\\
\end{tabular}
\end{center}
\vspace{-0.1in}
\caption{Parameter variation on the length of one video clip. }
\label{fig11}
\end{figure}

\subsubsection{Frequency ratio consistency loss $\mathcal L_\text{fr}$} \label{sec5.3.2}
Without $\mathcal L_\text{fr}$, the result also gets worse on both UBFC-rPPG and MMVS datasets. $\mathcal L_\text{fr}$ consists of two terms corresponding to the two positive signals, $\mathbf{y}^\mathbf{p}_1$ and $\mathbf{y}^\mathbf{p}_2$, which should ideally share the same frequency. Nonetheless, we argue that in practice both $\mathbf{y}^\mathbf{p}_1$ and $\mathbf{y}^\mathbf{p}_2$ are important in Eqn.\ref{equ4}. We offer a variant by keeping only the first term in Eqn.\ref{equ4} and name it as $\mathcal L_\text{fr} \rightarrow \mathcal L_\text{fr-single}$ in Table~\ref{table9}. This shows inferior performance compared to the original $\mathcal L_\text{fr}$ (Ours). Having both terms in Eqn.\ref{equ4} can implicitly pull the two positive signals close.

\subsubsection{Cross-video frequency agreement loss $\mathcal L_\text{fa}$} \label{sec5.3.3}
We conduct a similar experiment to that of $\mathcal L_\text{fr}$ by removing $\mathcal L_\text{fa}$, where a clear performance drop can be observed on both datasets. Next, we offer a variant of $\mathcal L_\text{fa}$ by keeping only its first term in Eqn.\ref{equ5}. We denote this by $\mathcal L_\text{fa} \rightarrow \mathcal L_\text{fa-single}$ in Table \ref{table9}. The result also gets worse, which further validates the importance of having both positive signals. 

\subsection{Temporally neighboring videos}\label{sec5.4} 
Given a short facial video, we cut it into $J+1$ clips, each with $T$ frames, and randomly select one as the main input $\mathbf{x^a}$. In this session, we study the attributes of these temporal neighbors.

\subsubsection{Number of video clips} \label{sec5.4.1}
We vary the number of video clips while fixing the length of {each} clip to 150. We vary $J$ from 1 to 6 in Fig.~\ref{fig12}. We only draw the MSE and RMSE results on HR estimation; the change for Pearson coefficient $r$ is insignificant. We can see the performance is stable on both datasets between $J = 1$ and 4. In general, we argue that $J$ should not be too big; otherwise, the signals from temporal neighbors would not be similar. Our default $J$ is 3.
%\ZJ{, which is equivalent to that of \cite{yu_physformer_2022}}.

\subsubsection{Length of one video clip} \label{sec5.4.2}
We also vary the length of each video clip $T$ from 50, 100, 150, 200, 250, and 300 in Fig.~\ref{fig11}. The number of video clips is fixed to 4 ($J =3$). 
%Note that the temporal neighbors of $\mathbf{x^a}$ share the same length with $\mathbf{x^a}$. We report the corresponding performance in Fig.~\ref{fig11}. 
We can observe that $T=150$ appears to be the best on UBFC-rPPG. The performance on MMVS is rather stable by varying $T$ between 100 and 200. Similar to $J$, $T$ should also not be too big such that neighboring signals remain to be similar. 

\subsubsection{Augmentation} \label{sec5.4.3}
We notice that for $\mathbf{x^a}$'s neighbors, they are only applied with the loss $\mathcal L_\text{fa}$, but are not augmented and applied with other losses ($\mathcal L_\text{fc}$, $\mathcal L_\text{fr}$, $\mathcal L_\text{vr}$). This is because the rPPG signals among neighboring video clips are very similar, there is no need to repeat the augmentation process for every one of them. To support our claim, we apply all loss terms to both $\mathbf {x^a}$ and its neighbors, and denote this by Ours-expand in Table \ref{table9}. Only slight improvement can be observed by comparing it to Ours. Considering the computation increase, we do not recommend it.

\begin{table}[!t]
\caption{Ablation study on spatial augmentation.}
\vspace{-0.1in}
\label{table10}
\begin{center}
\setlength{\tabcolsep}{1.5mm}
\begin{tabular}{c|ccc|ccc}
\toprule
\multirow{2}*{Method} &\multicolumn{3}{c|}{UBFC-rPPG}& \multicolumn{3}{c}{MMVS}\\ 

&MAE$\downarrow$  &RMSE$\downarrow$  &$r\uparrow$  & MAE$\downarrow$  &RMSE$\downarrow$  &$r\uparrow$ \\ 
\midrule
Ours ($p = 1$) &0.62  &1.00  &0.99  &2.95  &4.20  &0.94\\
Ours ($p = 2$) &0.58  &\textbf{0.94} &\textbf{0.99}  &\textbf{2.93} &4.16  &\textbf{0.94}\\
Ours ($p = 3$) &0.58  &0.95  &0.99  &2.93  &\textbf{4.15}  &0.94\\
Ours ($p = 4$) &\textbf{0.56}  &0.96  &0.99  &2.94  &4.16  &0.94\\
Ours-CJ &7.33  &8.97  &0.20  &8.29  &10.98  &0.47\\
Ours-SF &7.59  &9.55  &0.19  &8.33  &10.70  &0.46\\

\bottomrule
\end{tabular}
\end{center}
\vspace{-0.1in}
\end{table}

\subsection{Spatial Augmentation}\label{sec5.5}
We investigate the number of spatially augmented samples and the type of spatial augmentation. 

\subsubsection{The number of spatially augmented samples} \label{sec5.5.1}
We vary the number of positive (spatially augmented) samples from $p = 1$ to $p = 4$ in Table \ref{table10}.  When the number is 2, it appears to be a good trade-off between accuracy and computation. This is also our default setting. Notice that \cite{gideon_way_2021} only generates one positive sample by applying inverse transformation on the negative one, which is fundamentally different from our spatial augmentation. %In our way, if we also keep only one positive sample, the result  
%Ours-P1 is inferior to Ours-P2 because Ours-P2 can generate positive samples which have more spatial differences between each other. Ours-P3 and Ours-P4 do not show better performance instead brings in additional cost. Therefore, we select Ours-P2 by default
%and denote them as Ours-P1, Ours-P2, Ours-P3, Ours-P4. Note that in Ours-P1 we follow \cite{gideon_way_2021} to regard the anchor sample as another positive sample thus we can apply our frequency-inspired losses in this variant. The performance is reported .}

\subsubsection{The type of spatial augmentation} \label{sec5.5.2}
The spatial augmentation we adopt includes image rotation and flip. We choose them because they do not alter the colors across frames, thus do not affect the rPPG signals contained in videos. There exist many other types of spatial  augmentation, \eg color jittering and Sobel filtering. They are however not suitable for our task, as they clearly alter the color distribution across frames, so as to change the underlying signal frequencies in videos. To validate this, we offer two variants, denoted by Ours-CJ and Ours-SF, in Table \ref{table10}. For each variant, we generate two positive samples, one is the original input, the other is with color jittering or Sobel filtering. Both variants show very poor performance.

%Because these augmentations change the image color thus the rPPG signal is mixed with additional noises and is modified to of other frequencies. The augmented samples cannot be regarded as "positive" any more. Thus we only choose spatial augmentations which do not change the image color.}

%There is a high level of signal similarity in these clips thus regard all of them as anchor samples cannot improve the performance. Therefore, we only select one from them as the anchor sample and leverage the others for frequency supervision by the cross-video frequency agreement loss.
\section{Discussion}
In this section, we discuss the effect of makeup and the impact of ethnicity on our method, respectively. 

\begin{table}[!t]
\caption{Quantitative performance on subjects with or without makeup.}
\vspace{-0.1in}
\label{table12}
\begin{center}
\setlength{\tabcolsep}{1.5mm}
\begin{tabular}{c|ccc|ccc}
\toprule
\multirow{2}*{Method} & \multicolumn{3}{c|}{MMVS-w/ makeup}&\multicolumn{3}{c}{MMVS-w/o makeup}\\ 

&MAE$\downarrow$  &RMSE$\downarrow$  &$r\uparrow$  &MAE$\downarrow$  &RMSE$\downarrow$  &$r\uparrow$\\ 
\midrule
Gideon \etal \cite{gideon_way_2021} &3.90 &5.62 &0.92  &3.39  &4.69  &0.93\\
Ours &3.25  &4.46  &0.93  &2.90  &4.02  &0.94\\

\bottomrule
\end{tabular}
\end{center}
\vspace{-0.1in}
\end{table}

\begin{figure}[t]
\begin{center}
\begin{tabular}{cc}
\includegraphics[width=1.65in]{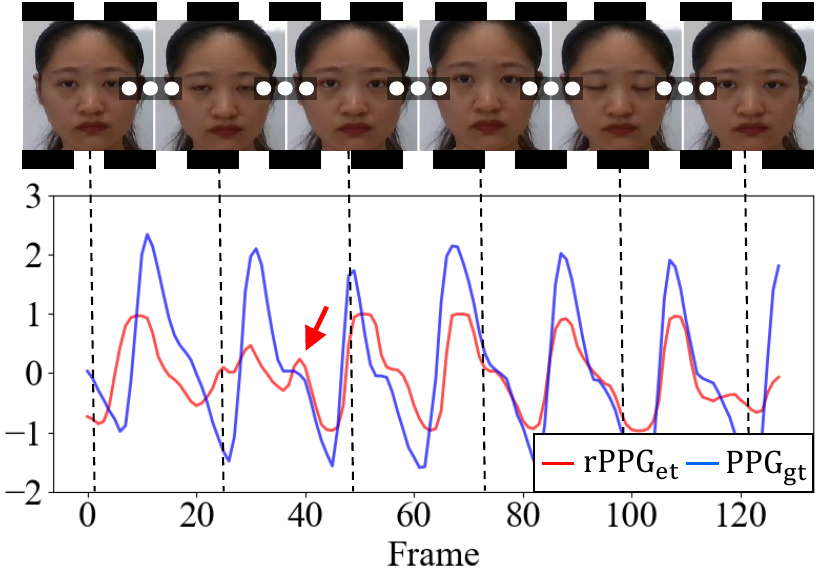} &
\includegraphics[width=1.65in]{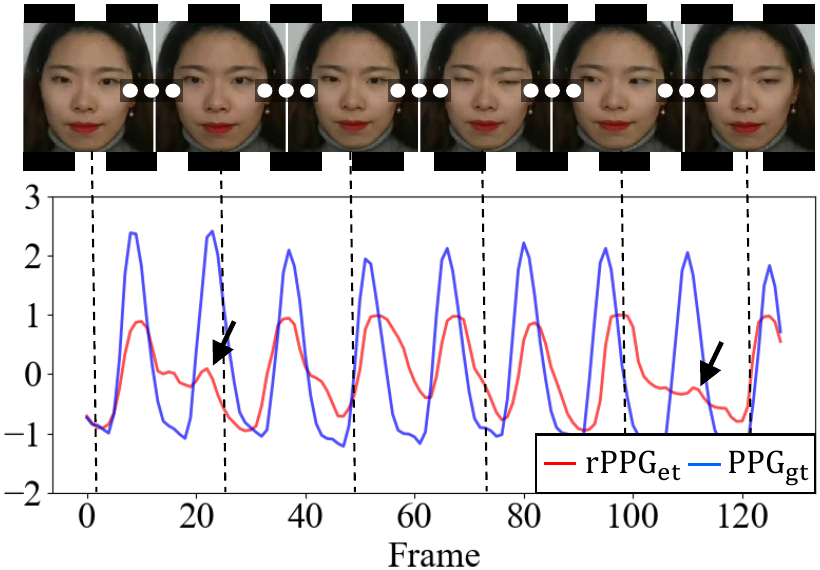} \\
\scriptsize (a) & \scriptsize (b)\\
\includegraphics[width=1.65in]{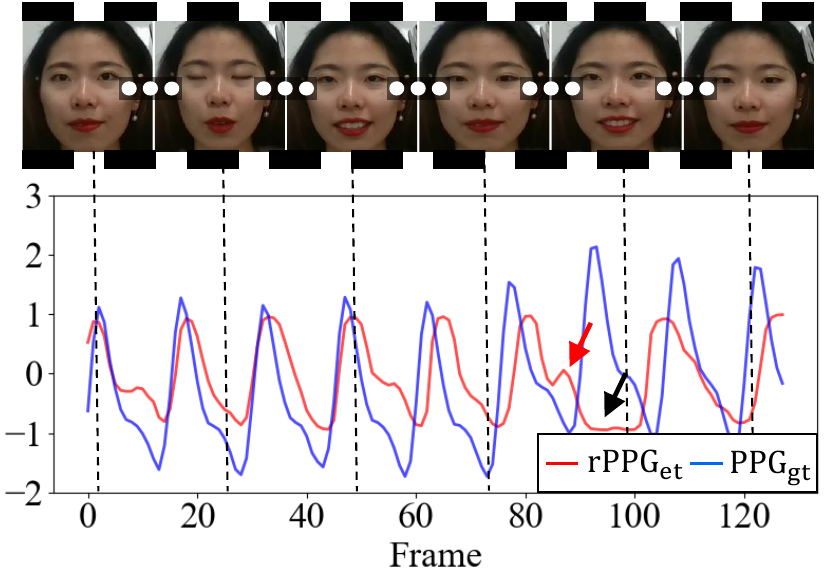} &
\includegraphics[width=1.65in]{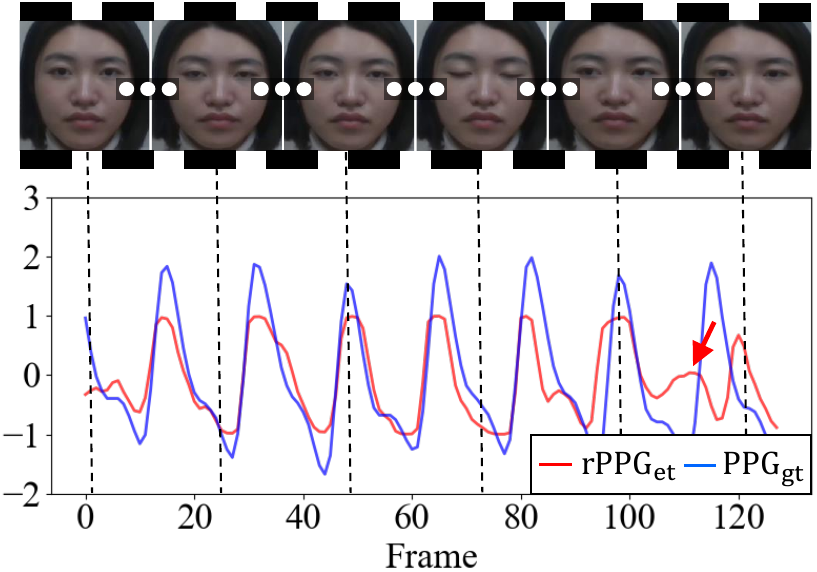} \\
\scriptsize (c) & \scriptsize (d)\\
\end{tabular}
\end{center}
\vspace{-0.1in}
\caption{Qualitative performance on subjects with makeup.}
\label{fig14}
\end{figure} 
\subsection{Effect of makeup}
We first discuss the effect of makeup on the model performance. Among five datasets, MMVS contains subjects with makeup. Therefore, we split the MMVS into two subsets: MMVS-w/ makeup and MMVS-w/o makeup.
% , and evaluate our method on these two subsets. 
Each subset contains a training set and a test set, which are respectively leveraged for model training and evaluation.
HR estimation results are shown in Table \ref{table12}. It can be seen that the results on MMVS-w/o makeup are better than those on MMVS-w/ makeup: \eg the MAE is 2.90 on the former and 3.25 on the latter. Also, compared to the state of the art \cite{gideon_way_2021}, which reflects a difference of 0.51 MAE between the two subsets, the influence of makeup on our method is smaller (\ie 0.35 MAE difference between the two subsets). Fig.~\ref{fig14} further shows four estimated rPPG signals and their corresponding ground truth signals from subjects with makeup. We can see that some signal peaks are not well reconstructed (black arrows) while some are mistakenly reconstructed (red arrows). Makeup can cover the periodic color changes on faces, making it difficult to capture physiological clues for accurate rPPG estimation.

\begin{table*}[!t]
\caption{Quantitative performance on subjects of different ethnicities.}
\vspace{-0.1in}
\label{table13}
\begin{center}
\setlength{\tabcolsep}{1.5mm}
\begin{tabular}{c|ccc|ccc|ccc}
\toprule
\multirow{2}*{Method} & \multicolumn{3}{c|}{BP4D-Caucasian}&\multicolumn{3}{c|}{BP4D-African}&\multicolumn{3}{c}{BP4D-Asian}\\ 

&MAE$\downarrow$  &RMSE$\downarrow$  &$r\uparrow$  &MAE$\downarrow$  &RMSE$\downarrow$  &$r\uparrow$&MAE$\downarrow$  &RMSE$\downarrow$  &$r\uparrow$\\ 
\midrule
Gideon \etal \cite{gideon_way_2021} &3.68 &4.96 &0.88  &5.15  &7.00  &0.82&3.22  &4.40  &0.92\\
Ours &2.89  &4.13  &0.93  &4.69  &6.21  &0.85 &2.54  &3.72  &0.96\\

\bottomrule
\end{tabular}
\end{center}
\vspace{-0.1in}
\end{table*}

\begin{figure}[t]
\begin{center}
\begin{tabular}{cc}
\includegraphics[width=1.65in]{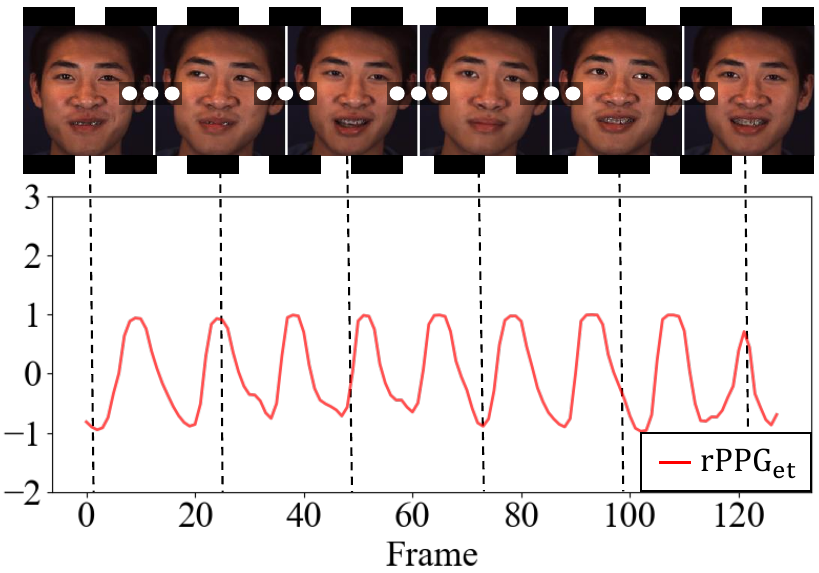} &
\includegraphics[width=1.65in]{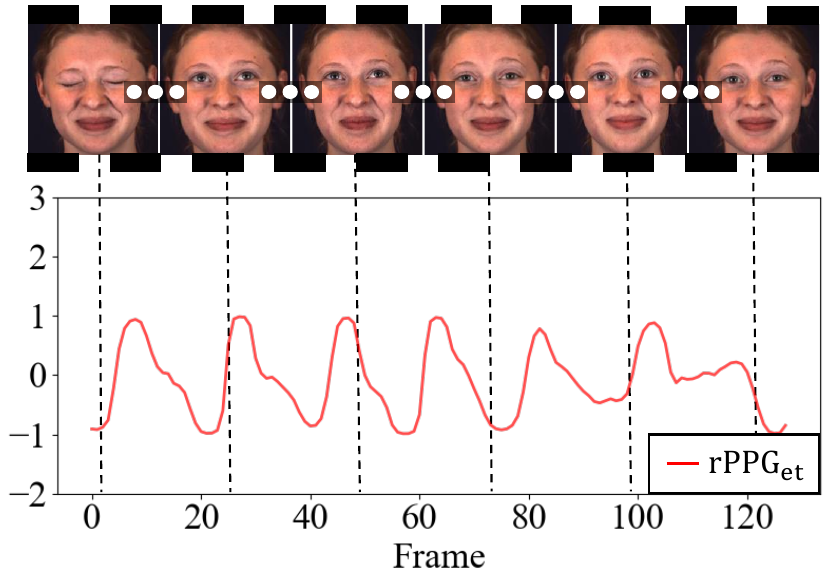} \\
\scriptsize (a) Asian & \scriptsize (b) Caucasian\\
\includegraphics[width=1.65in]{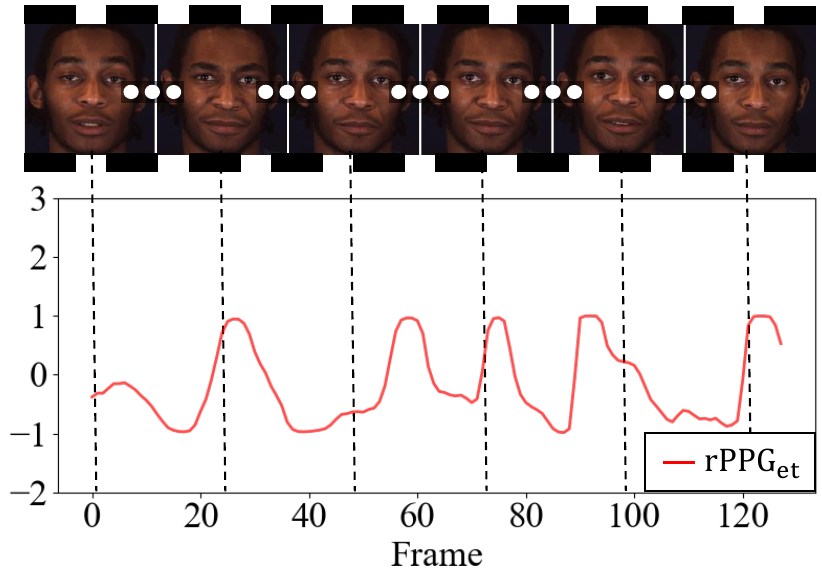} &
\includegraphics[width=1.65in]{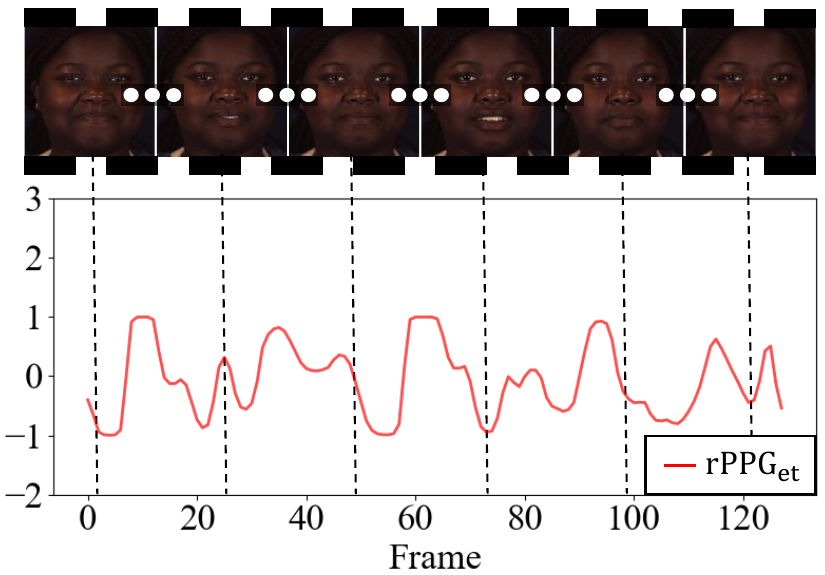} \\
\scriptsize (c) African & \scriptsize (d) African\\
\end{tabular}
\end{center}
\vspace{-0.1in}
\caption{Qualitative performance on subjects of different ethnicities.}
\label{fig15}
\end{figure} 

\subsection{Impact of ethnicity}\label{sec6.2}
Next, we evaluate the model performance on subjects of different ethnicities. Among the five datasets, BP4D+ includes a wide coverage of ethnicities from Caucasian, African and Asian (East-Asian and Middle-East-Asian). 
We split the BP4D+ dataset into three subsets according to the subject ethnicities, \ie BP4D-Caucasian, BP4D-African and BP4D-Asian. Each subset is divided into the training set and the test set. We train our method on the training set of each subset and evaluate it on the corresponding test set. HR estimation results on three test sets are shown in Table \ref{table13}. Fig.~\ref{fig15} shows four concrete examples. From Table \ref{table13} we can observe that both our approach and \cite{gideon_way_2021} perform the best on BP4D-Asian, the second best on BP4D-Caucasian and the worst on BP4D-African. Fig.~\ref{fig15}(a) shows satisfactory signal periodicity on an Asian subject, while the estimated signals from African subjects in Fig.~\ref{fig15}(c) and Fig.~\ref{fig15}(d) are of low quality, especially their interbeat intervals vary significantly; we cannot observe the periodicity or quasi-periodicity from their waveforms. Note that BP4D+ does not include ground truth PPG signals but only the ground truth of HR.
 
According to \cite{mcduff_camera_2023}, most RGB cameras are optimized to capture light skin tones more effectively than dark skin tones. Typically, a camera's sensitivity is the greatest in the middle of the (0, 255) pixel range. That means dark skin tones of African subjects are likely to saturate the pixels, which would cause the skin color change by the physiological signal variation to be suppressed. Both our approach and \cite{gideon_way_2021} cannot obtain sufficient physiological clues for generating periodic rPPG signals on African subjects. Our finding is consistent with that reported in \cite{mcduff_camera_2023}.

Given that BP4D+ is an ethnicity-imbalanced dataset (see Sec.~\ref{sec4.1}), we further randomly sample 15 subjects of each ethnicity to construct the subsets BP4D-Caucasian, BP4D-Asian and BP4D-African again with balanced distribution. We train and test the three ethnicity-specific models on the three subsets respectively. According to our experiments on the HR estimation, the performance on BP4D-Caucasian and BP4D-Asian is still better than that on BP4D-African. For example, the MAE on BP4D-Caucasian and BP4D-Asian is 3.13 and 2.81, respectively; while on BP4D-African is 4.69. These results further validate our argument: the saturation of skin pixels of African subjects can suppress the pulsation-induced skin color change.
\section{Conclusion}

In this paper, we propose a new frequency-inspired self-supervised framework for facial video-based remote physiological measurement. Our approach consists of three key stages: data augmentation, signal extraction and network optimization. For data augmentation, we randomly select a video clip from a given facial video as the main input. We introduce the LFA module to generate sufficient negative samples from it. These negative samples contain rPPG signals with different frequencies to the input. Meanwhile, we also apply spatial augmentation on the input to obtain positive samples that share the same rPPG signal frequency to that of the input. For signal extraction, we design the REA module to extract complementary physiological clues from different face regions and aggregate them for an accurate rPPG estimation. In the REA module, a region-attention block is devised to focus the estimation on pulsation-sensitive skins and a spatio-temporal gating net is devised to combine {temporal signals over different spatial regions.}
%while suppressing noises from irrelevant backgrounds and pulsation-insensitive skins. 
Last, for the network optimization, we propose a series of frequency-inspired losses to optimize estimated rPPG signals from the input's augmentations and neighbors. Experiments on five datasets show the effectiveness and superiority of our method over state of the art.
%\ZJ{In the future work, we would like to improve the framework to learn contrastive frequency information among videos from different subjects for boosting the rPPG estimation performance.}
\section*{Acknowledgments}

This work was supported by the National Natural Science Foundation of China [No. 72293581, 72293580, 72188101 and 62103122], and the Fundamental Research Funds for the Central Universities.

\bibliographystyle{IEEEtran}   
\bibliography{IEEEabrv,collection.bib}

\begin{IEEEbiography}
[{\includegraphics[width=1in,height=1.25in,clip,keepaspectratio]{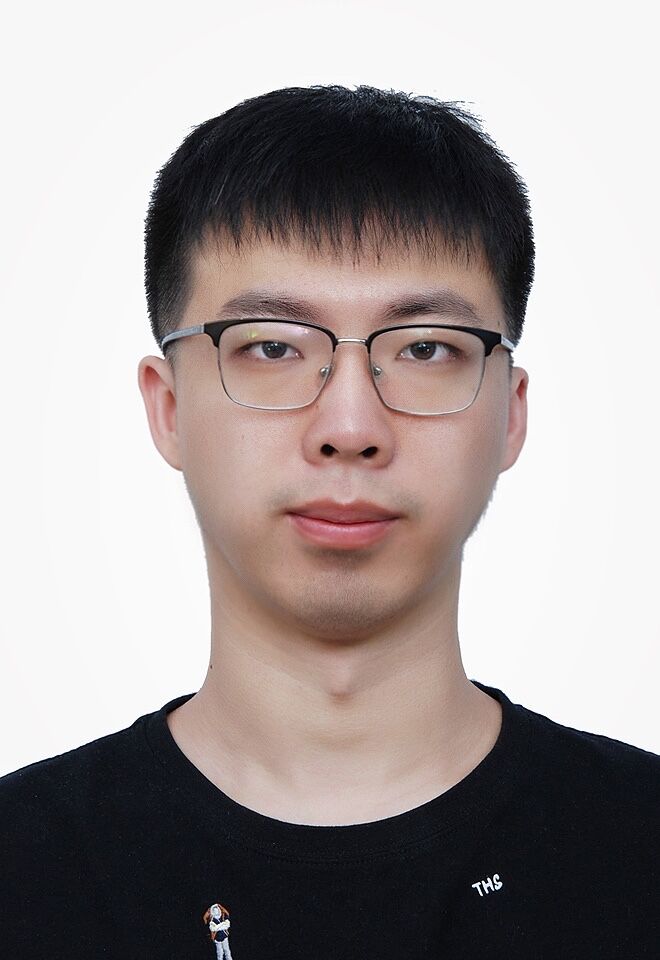}}]
{Zijie Yue} received his B.E. and Ph.D. degrees from Hefei University of Technology in 2017 and 2023, respectively. He is currently a postdoctoral researcher at Tongji University. His current research is in the area of remote physiological measurement, computer vision, and machine learning.
\end{IEEEbiography}
\begin{IEEEbiography}
[{\includegraphics[width=1in,height=1.25in,clip,keepaspectratio]{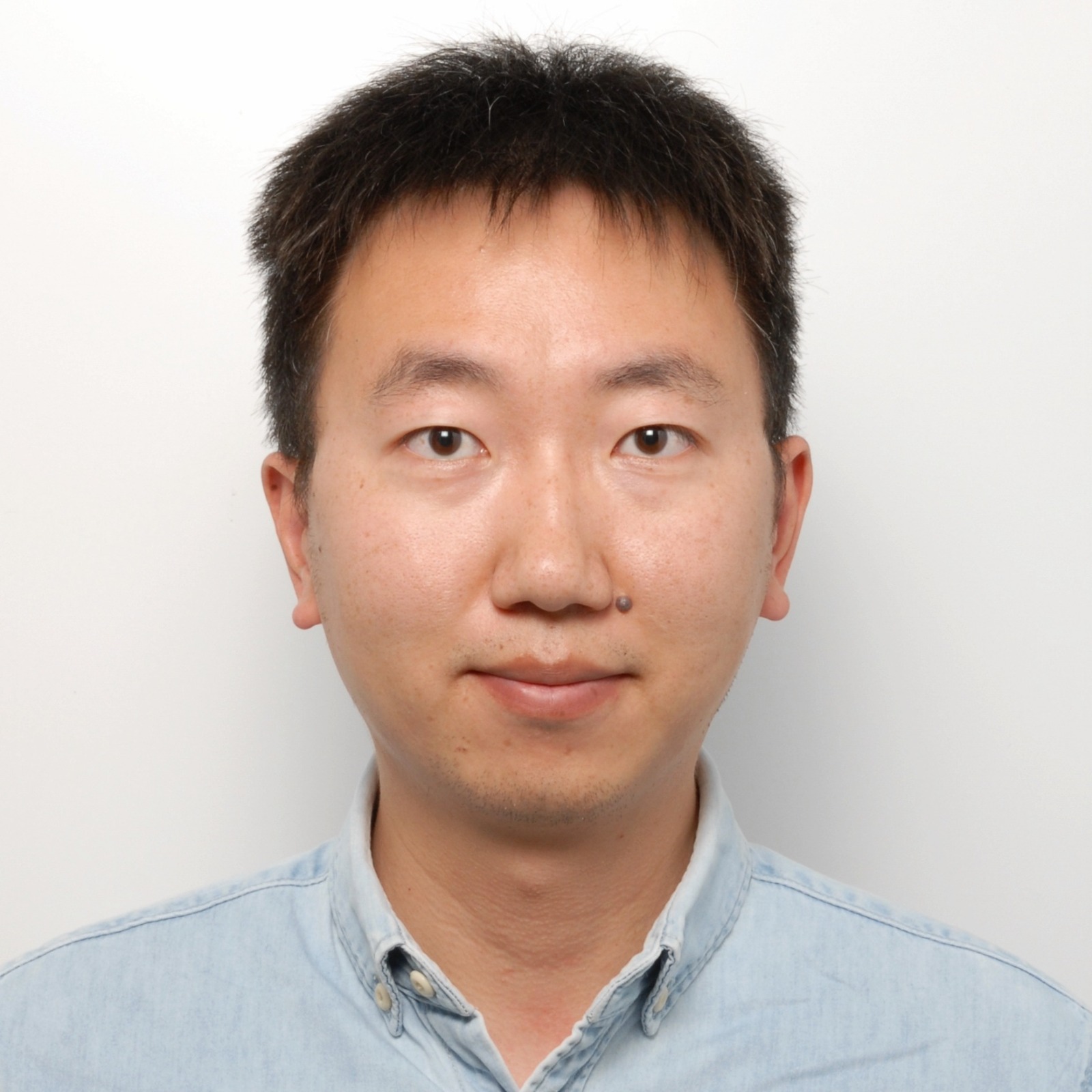}}]
{Miaojing Shi} (Senior Member, IEEE) received the Ph.D. degree from Peking University in 2015. He also engaged with a joint Ph.D. program with the University of Oxford and INRIA Rennes for a year. He held a postdoctoral position at the University of Edinburgh and was a Research Scientist at INRIA Rennes. Between 2020 and 2022, he has been a Lecturer with the Department of Informatics, King's College London. Since 2023, he becomes a Full Professor at Tongji University and a visiting Senior Lecturer at King’s. He has authored or coauthored over 50 publications. His current research focus is on visual learning and understanding, with multi-task and limited supervision.
\end{IEEEbiography}
\begin{IEEEbiography}
[{\includegraphics[width=1in,height=1.25in,clip,keepaspectratio]{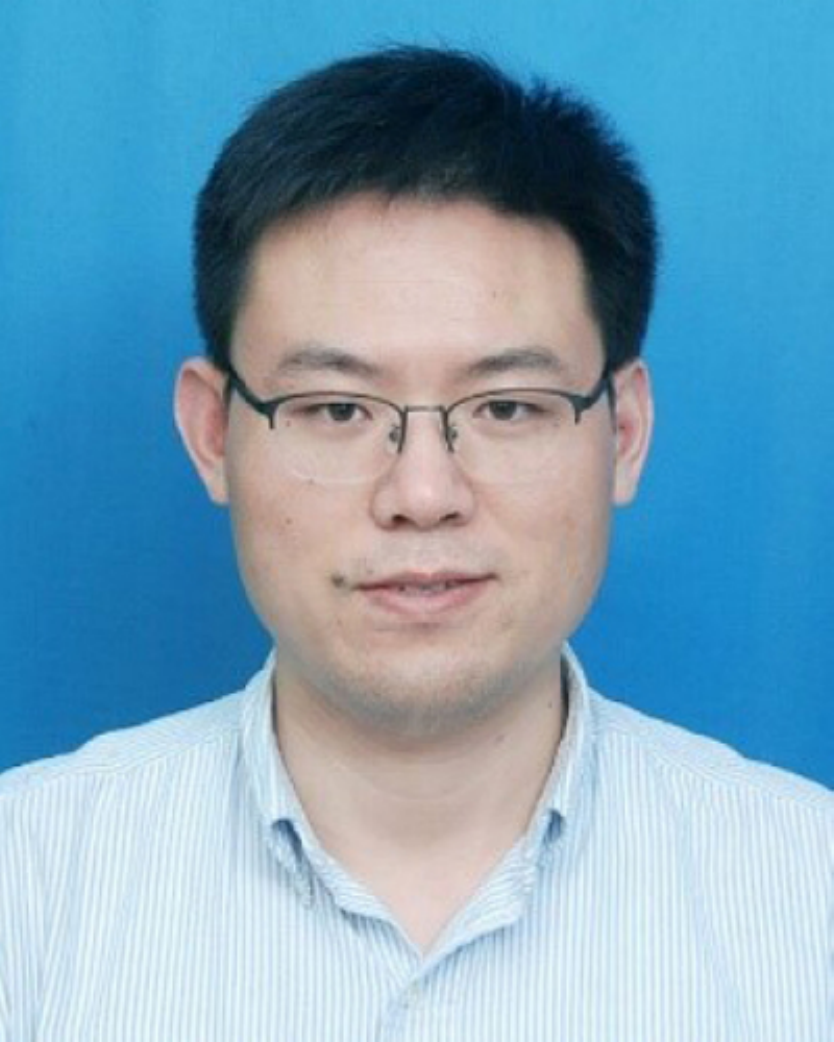}}]
{Shuai Ding} received his Ph.D. degree in management science and engineering from Hefei University of Technology, Hefei, China, in 2011. He has been a Visiting Scholar with the University of Pittsburgh, Pittsburgh, PA, USA. He is currently a Professor with the School of Management, Hefei University of Technology. His research interests include smart healthcare, medical artificial intelligence, and management information systems.
\end{IEEEbiography}
\end{document}